\documentclass[sigconf]{acmart}

\usepackage{amsmath}
\usepackage{wrapfig}
\usepackage{graphicx}
\usepackage{subfig}
\usepackage{caption}
\usepackage{amsfonts}       
\usepackage{hyperref}
\usepackage{booktabs}       
\usepackage{siunitx}
\usepackage{url}
\usepackage{cleveref}
\usepackage{wrapfig}
\usepackage[T1]{fontenc}

\definecolor{light-gray}{gray}{0.5}
\definecolor{dark-gray}{gray}{0.2}

\crefname{chapter}{\S}{\S\S}
\crefname{section}{\S}{\S\S}

\definecolor{custom-purple}{RGB}{138,43,226}
\definecolor{custom-green}{RGB}{102,205,170}
\definecolor{custom-orange}{RGB}{255,140,0}

\copyrightyear{2023}
\acmYear{2023}
\setcopyright{acmlicensed}
\acmConference[FAccT '23]{2023 ACM Conference on Fairness, Accountability, and Transparency}{June 12--15, 2023}{Chicago, IL, USA}
\acmBooktitle{2023 ACM Conference on Fairness, Accountability, and Transparency (FAccT '23), June 12--15, 2023, Chicago, IL, USA}
\acmPrice{15.00}
\acmDOI{10.1145/3593013.3594003}
\acmISBN{979-8-4007-0192-4/23/06}

\begin{document}

\title{Simplicity Bias Leads to Amplified Performance Disparities}


\author{Samuel J.~Bell}
\email{sjbell@meta.com}
\orcid{0000-0002-9437-5449}
\affiliation{%
  \institution{FAIR, Meta AI}
  \city{Paris}
  \country{France}}

\author{Levent Sagun}
\email{leventsagun@meta.com}
\orcid{0000-0001-5403-4124}
\affiliation{%
  \institution{FAIR, Meta AI}
  \city{Paris}
  \country{France}}

\renewcommand{\shortauthors}{Bell and Sagun}



\begin{abstract}

\textbf{Which parts of a dataset will a given model find difficult?}
Recent work has shown that SGD-trained models have a bias towards simplicity, leading them to prioritize learning a majority class, or to rely upon harmful spurious correlations. 
Here, we show that the preference for `easy' runs far deeper: A model may prioritize any class or group of the dataset that it finds simple---at the expense of what it finds complex---as measured by performance difference on the test set.
When subsets with different levels of complexity align with demographic groups, we term this \emph{difficulty disparity}, a phenomenon that occurs even with balanced datasets that lack group/label associations.
We show how difficulty disparity is a model-dependent quantity, and is further amplified in commonly-used models as selected by typical average performance scores.
We quantify an \emph{amplification factor} across a range of settings in order to compare disparity of different models on a fixed dataset.
Finally, we present two real-world examples of difficulty amplification in action, resulting in worse-than-expected performance disparities between groups even when using a balanced dataset.
The existence of such disparities in balanced datasets demonstrates that merely balancing sample sizes of groups is not sufficient to ensure unbiased performance.
We hope this work presents a step towards measurable understanding of the role of model bias as it interacts with the structure of data, and call for additional model-dependent mitigation methods to be deployed alongside dataset audits.

\end{abstract}



\begin{teaserfigure}
    \centering
    \subfloat[Example: shape classification]{
        \raisebox{-0.35em}{\includegraphics[trim={2cm 0.75cm 2cm 2.5cm},clip,width=0.275\textwidth]{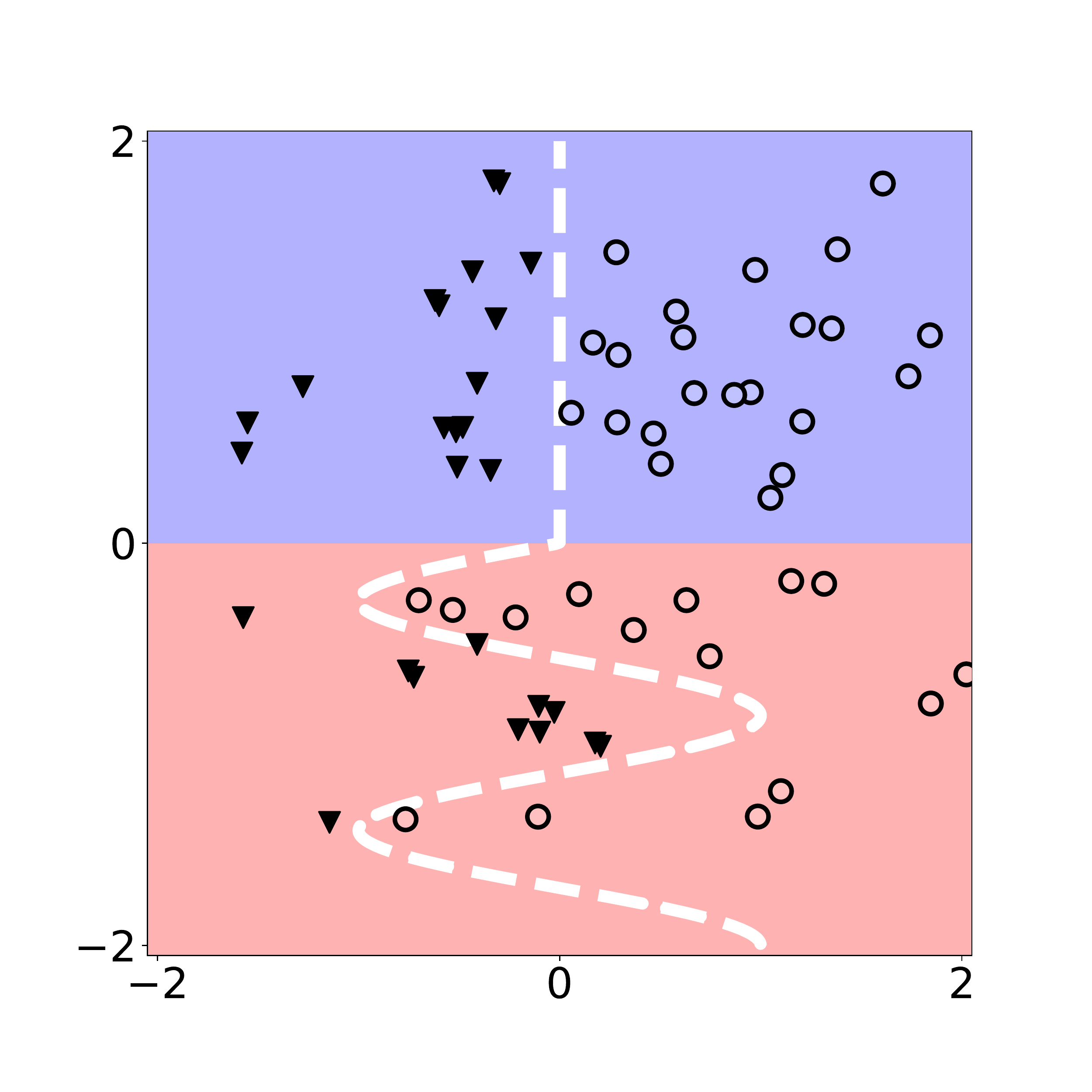}}
    }
    \subfloat[Train each group separately]{
        \includegraphics[trim={0.5cm 0.5cm 0.5cm 0.5cm},clip,width=0.28\textwidth]{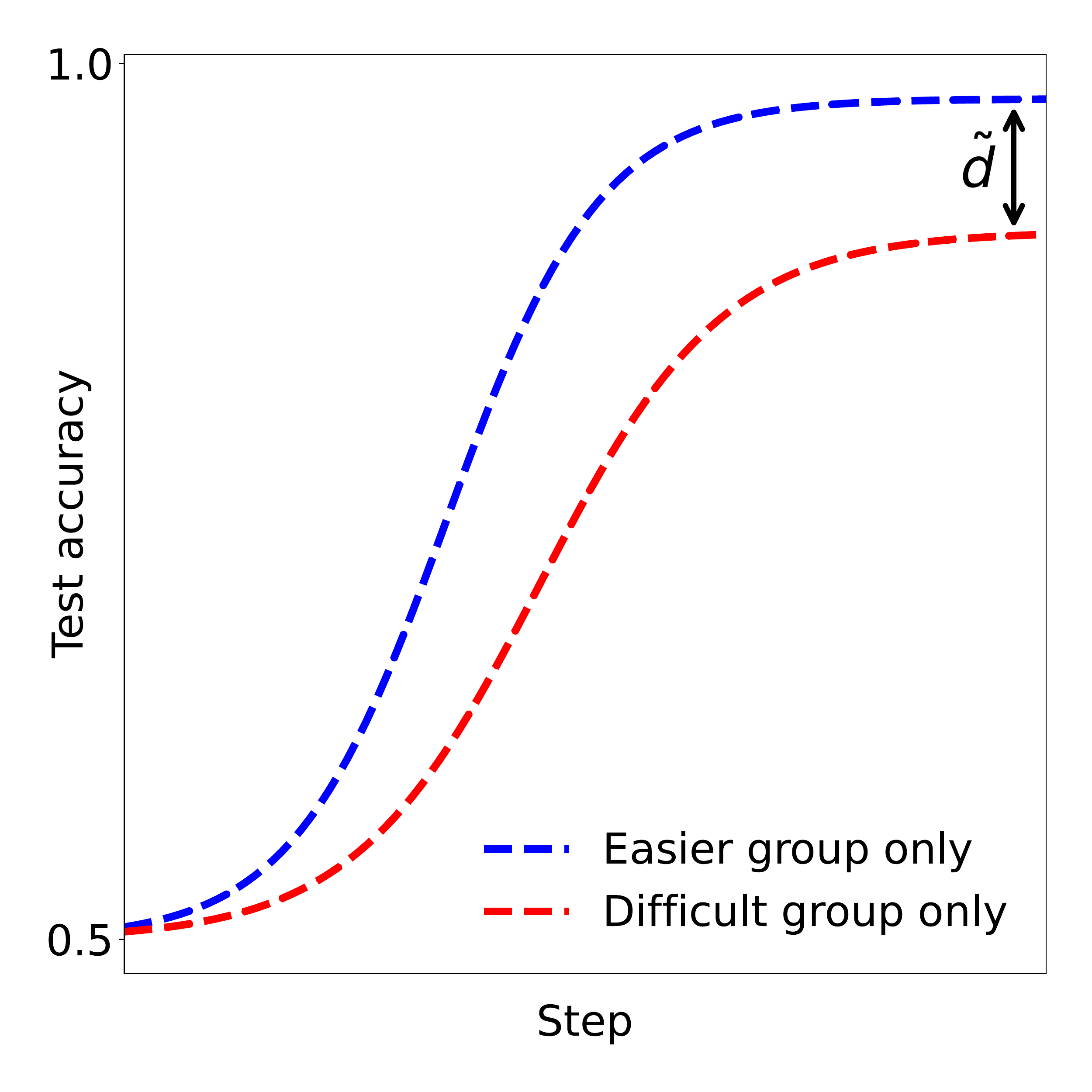}
    }
    \subfloat[Train both groups together]{
        \includegraphics[trim={0.5cm 0.5cm 0.5cm 0.5cm},clip,width=0.28\textwidth]{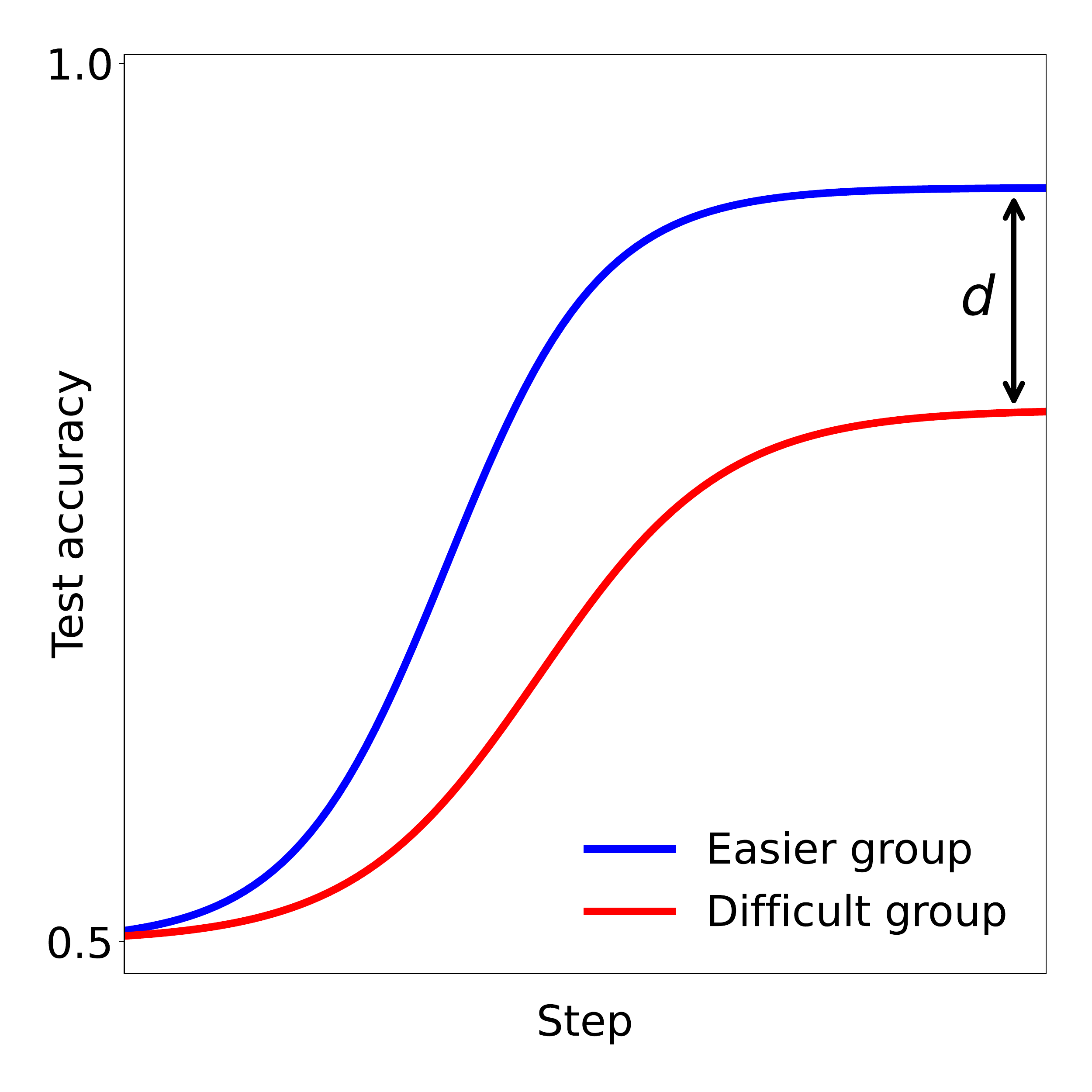}
    }
    \caption{\textbf{Sketch of difficulty amplification:} \textbf{(a)}~Consider a binary classification of hollow circles vs.~filled triangles where a potential decision boundary is indicated via the white dashed line. Above \(y=0\) (\textcolor{blue}{blue}) we have a simple group which is linearly separable. Below \(y=0\) (\textcolor{red}{red}) we have a more complex group with a non-linear decision boundary. \textbf{(b)}~Illustration of test accuracy when training on the simple group only (\textcolor{blue}{blue dashed}) and the complex group only (\textcolor{red}{red dashed}). As expected we obtain better accuracy on the simple group. \textbf{(c)}~However, when training on both groups at once (solid), the model exacerbates the difference: the observed accuracy disparity $d$ (vertical arrow in (c)) exceeds the estimated accuracy disparity from individual group training $\tilde{d}$ (vertical arrow in (b)). When $d>\tilde{d}$, we call this \textbf{difficulty amplification}.}
    \label{fig:sketch-alt}
\end{teaserfigure}

\received{6 February 2023}
\received[accepted]{7 April 2023}

\maketitle

\section{Introduction}

Without actually training a model, understanding what the model will find challenging is far from trivial.
It is widely known that a certain dataset may be hard for one model but not for another \citep{Wolpert1997}.
For a given model, two classes may be easily separable, while for another they may be hard to distinguish.
As such, it follows naturally that `difficulty' is a function of both \emph{data and model}, such that we can't properly account for difficulty by analyzing the dataset alone.

In the context of machine learning fairness, for a given task and dataset, a model may find one social group more difficult than another, leading to disparate impact \citep{barocas2016big}.
For example, \citeauthor{Buolamwini2018}'s \cite{Buolamwini2018} audit of commercial image recognition systems finds they exhibit worse accuracy for darker-skinned women than for any other group.
Typically, such accuracy disparity is attributed either to under-representation of certain groups, or to spurious correlations between group information and target variable.
In this work, we further show that---even with perfectly balanced data and in the absence of associations between group labels and class labels---trained models can and do find certain groups harder than others.
Crucially, group difficulty is hard to ascertain during dataset audit, providing key evidence for the necessity of a complementary post-training model audit. 

Given the existence of model-specific difficulty differences, we consider the role of the model itself.
It is well known that many contemporary models are biased towards learning simple functions \citep{Arpit2017, Kalimeris2019, Rahaman2019, Valle-Perez2019, Shah2020}, a phenomenon recently linked to an over-reliance on spurious correlations \citep{Shah2020, Sagawa2020}.
Here, we push this line of inquiry further, and show that simplicity bias can be harmful in an entirely different way: When a model finds one group easier than another (even if sample sizes of each group are balanced), it will prioritize the easy group at the expense of the difficult, resulting in a greater performance disparity when compared to training each group separately.
We refer to this phenomenon as \emph{difficulty amplification} (see illustration in \cref{fig:sketch-alt}).
Through experiments using balanced datasets (with identical label distributions between groups), we show that between-group difficulty amplification is distinct and separate from a majority group preference, and from learning simple ``shortcuts'' \citep{Shah2020, Geirhos2020}.
Our experiments further reveal that difficulty amplification is sensitive to model architecture, training time, and parameter count.
Seemingly innocuous design decisions, such as whether to use early stopping, affect both amplification factor and the resulting performance disparity.

We demonstrate the substantial and heterogeneous impact of difficulty amplification via two case studies. 
First, we observe a worse-than-expected performance disparity between race-annotation groups on an age classification task using FairFace \citep{Karkkainen2021}.
Second, we show a worse-than-expected performance disparity between household income quartiles during an object classification task using Dollar Street \citep{DollarStreet}.
We also explore the potential impact of additional data collection and oversampling as they are common strategies relied upon for mitigating disparate outcomes. 
We present preliminary yet positive results highlighting the need for bias mitigation strategies that focus not only on the choices pertaining the core components of training: \textbf{dataset}, \textbf{architecture}, and \textbf{algorithm}; but also their interactions with each other.

\hfill\break
\noindent To summarize, we highlight the following conceptual contributions in this work:
\begin{enumerate}
    \item \textbf{difficulty disparity}, a pervasive phenomenon that persists in models even after dataset audits,
    \item \textbf{difficulty amplification factor}, quantifying how much a model exacerbates difficulty disparity, and 
    \item \textbf{empirical evaluation} showing the role of model architecture, training time, and parameter count.
\end{enumerate}
\noindent Building on our findings of deep data---model interdependence, we argue in favor of model-specific fairness audits by:
\begin{enumerate}
    \item demonstrating the substantial impact of difficulty amplification on \textbf{two real-world case studies}, and 
    \item exploring the impact of two common \textbf{mitigation strategies}---additional data collection and oversampling---both of which are performed \textit{after} having identified model-dependent performance disparities.
\end{enumerate}

\section{Background and related work}\label{sec:background}

\subsection{Biased datasets and biased models}

\emph{Bias} in ML systems arises from many sources.
At the most basic level, a dataset itself is biased if certain groups are under-represented \citep{Stock2018, Hendricks2018, Yang2020, Menon2021}.
Proposals to rectify under-representation include actively collecting more data for certain groups \citep{Dutta2020}, under/oversampling or reweighting during training \citep{Byrd2019, Sagawa2020, Idrissi2022, Arjovsky2022}, and optimizing for worst-group (as opposed to average) accuracy \citep{Sagawa2019}.
Recent work has suggested fine-tuning on an explicitly balanced set \citep{Kirichenko2022}.

Alternatively, datasets can reinforce harmful associations \citep{Goyal2022a}, both due to sampling error and by inadvertently capturing an undesirable association that is present in society.
Bucketed as \textit{spurious correlations} \citep{Muthukumar2018, Wang2019, Sagawa2020} or \textit{shortcuts} \citep{Geirhos2020}, a large body of fairness work seeks to train models that learn some true function invariant to the spuriously correlated feature.
Both under-representation and spurious correlations dominate the fairness literature landscape, though in both cases the onus is squarely on the data.
Like our work, \citeauthor{Wang2019}~\cite{Wang2019} have argued that balancing a dataset is not sufficient to preclude biased model performance, attributing the resulting bias to hidden (i.e., unlabeled) spurious correlations or confounders in the dataset.
In contrast, our work differs through its focus on an entirely distinct contributing factor: the relative complexity difference of the groups, which we explore using experiments manipulating the difficulty gap. 

\begin{figure*}[t]
    \centering
    \subfloat[]{
        \includegraphics[trim={0 0 0 0},clip,width=0.32\textwidth]{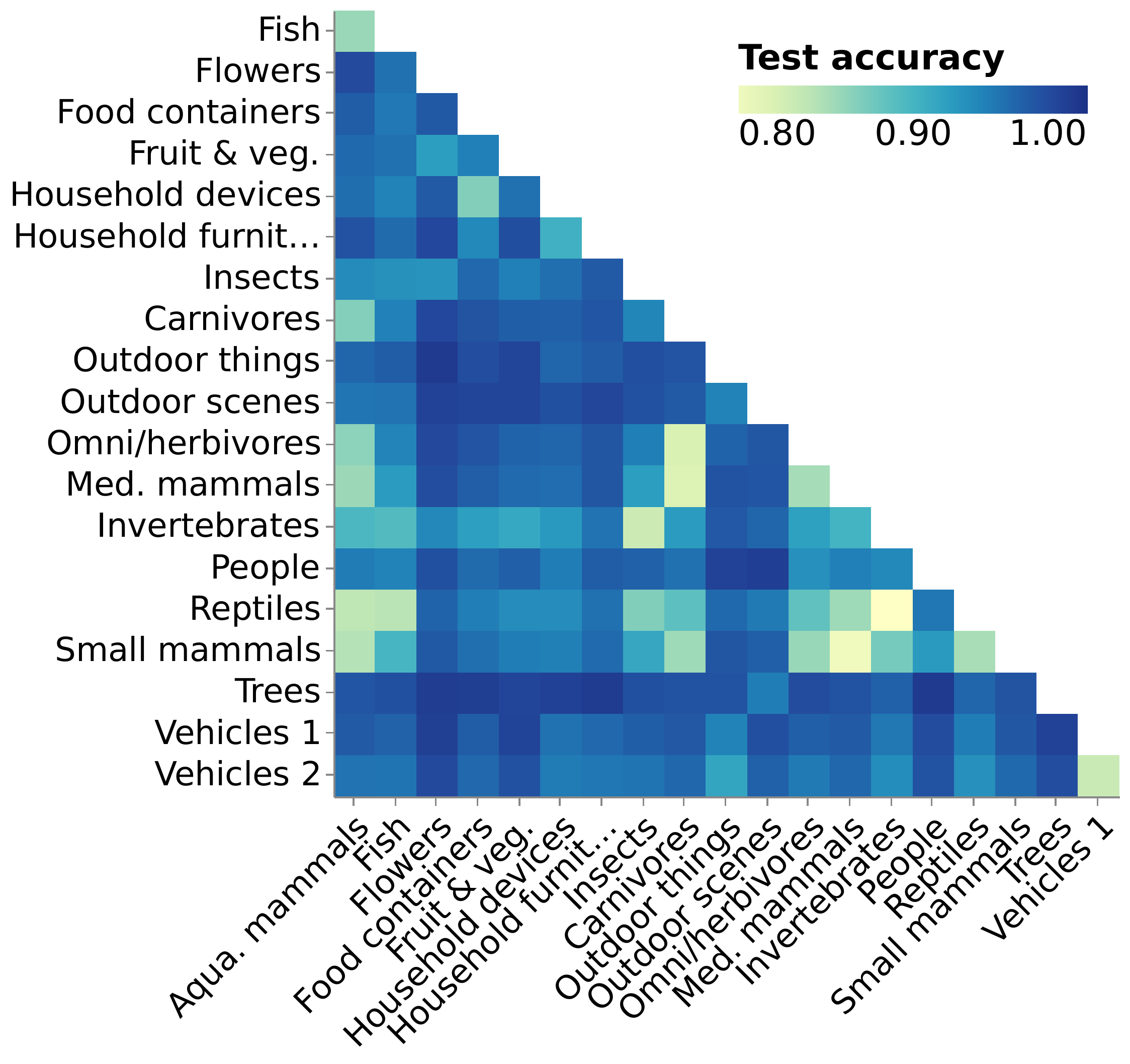}
    }
    \subfloat[]{
        \raisebox{1em}{\includegraphics[trim={0 0 0 0},clip,width=0.19\textwidth]{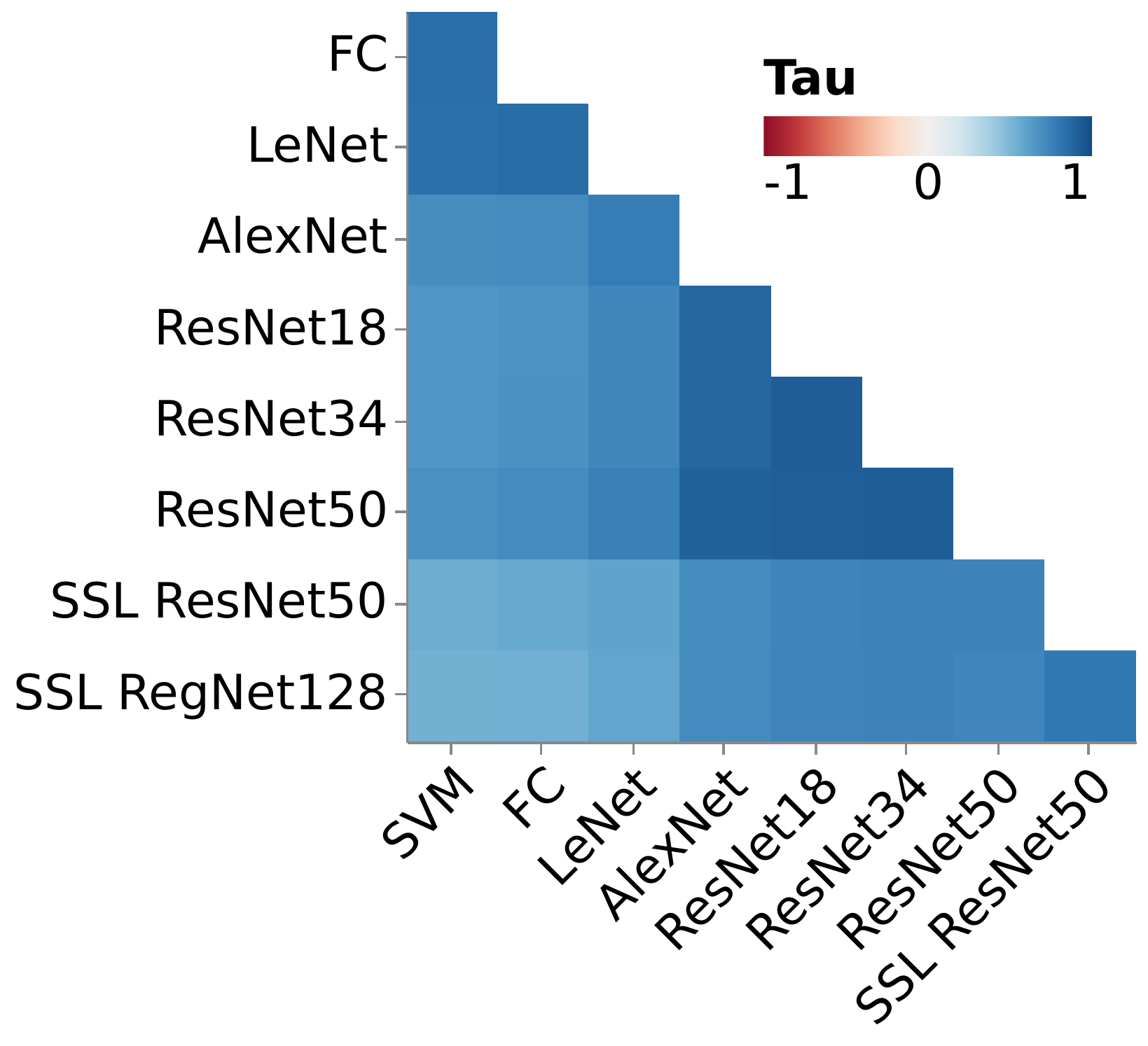}}
    }
    \subfloat[]{
        \includegraphics[trim={0 0 0 0},clip,width=0.32\textwidth]{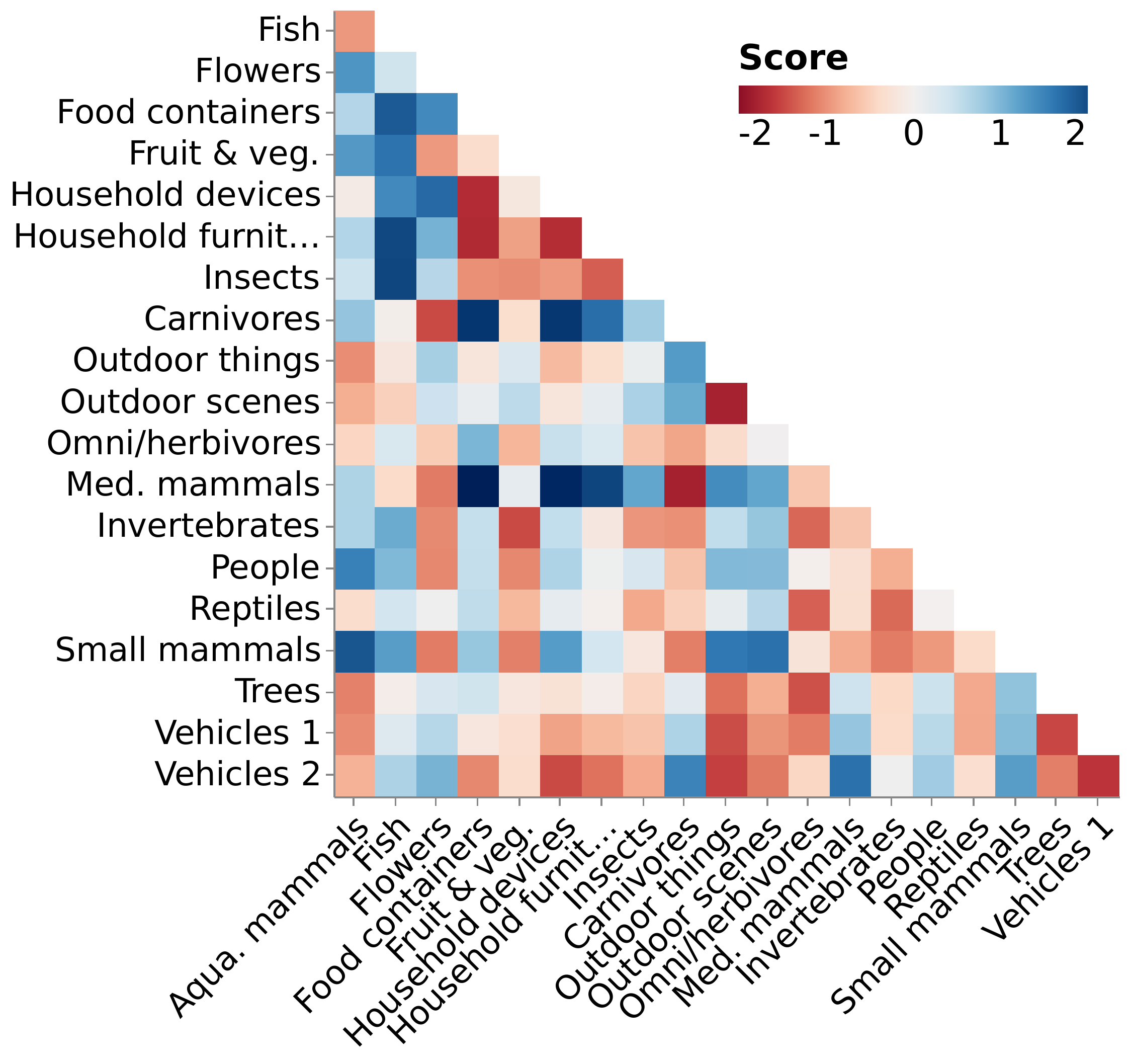}
    }
    \caption{\textbf{(a)}~ResNet-18 finds certain class pairs of CIFAR-100 more difficult than others, according to binary test accuracy. \textbf{(b)}~Rank ordering of difficulty varies across models, measured here by Kendall's \(\tau\). \textbf{(c)}~1D PLS projection of binary accuracies onto cosine distance between class mean vectors. Red cells---pairs where increasing inter-class distance decreases accuracy---demonstrate \textbf{there is no clear way to identify what a model will find difficult without training.}
    }\label{fig:difficulty}
\end{figure*}

\subsection{Beyond spurious correlations}\label{sec:background-beyond-spurious}

There is increasing focus on the model itself, independent of the role of data \citep{Hooker2021}, of which \emph{bias amplification} is a relevant example \citep{Zhao2017, Wang2021}.
Here, a small correlation in the training set is amplified into a larger correlation at test time.
In empirical experiments evaluating bias amplification, \citeauthor{Hall2022}~\cite{Hall2022} suggest that if group membership is easier to identify than class membership, models prefer to use the spurious correlation.
They also  report no bias amplification in the case where there is no spurious correlation present in the data: this is expected given the definition of amplification involves a multiplication of an existing data bias. 

Other subtle biases have been identified in the balanced data setting.
\citeauthor{Leino2018}~\cite{Leino2018} show that models trained with SGD overly rely upon moderately spuriously-correlated features if they are sufficiently numerous relative to the size of training set.
\citeauthor{Khani20a}~\cite{Khani20a} find that adding feature noise equally across groups induces disparity, a fact that can also be attributed to the relative difficulty of group information versus the desired target.
\citeauthor{Khani2021}~\cite{Khani2021} find removing spurious features can disproportionately lower performance on certain groups, and argue (as do we) that a balanced dataset is not a sufficient guard against biased performance.
\citeauthor{Mannelli2022}~\cite{Mannelli2022} use teacher-student networks to show subtle properties such as differences in group distance from the overall mean and differences in group variance are sufficient to induce biased outcomes in the absence of spurious correlations. 
Each of these works adds credence to a central notion of our work: that dataset bias is hard to identify, difficult to remove, and yet doing so is not necessarily sufficient to reduce model bias.

\subsection{Inductive bias towards simplicity}

Numerous recent works have identified the tendency for SGD-trained models to prioritize simple data points during training, resulting in simple functions being learned before more complex ones \citep{Arpit2017, Kalimeris2019, Valle-Perez2019}.
\citeauthor{Jo2017}~\cite{Jo2017} show that convolutional networks are overly dependent on surface-level statistical properties of images, such that applying a Fourier filter to the training set is sufficient to radically degrade test performance.
\citeauthor{Rahaman2019}~\cite{Rahaman2019} also show models are biased towards low-frequency functions, learning these simpler functions before more complex, higher-frequency examples.
Though often framed as a positive, allowing neural networks to learn functions that generalize well by applying Occam's razor, \citeauthor{Shah2020}~\cite{Shah2020} cite this simplicity bias as potentially harmful, at root causing both vulnerability to adversarial attacks and over-reliance on spurious correlations.
Similarly, \citeauthor{Dagaev2021}~\cite{Dagaev2021} reserve as much skepticism for overly simple solutions as for overly complicated, arguing that excessively simple solutions are likely to rely on potentially harmful ``shortcuts'' \citep{Geirhos2020}.
\citeauthor{Sagawa2020}~\cite{Sagawa2020} find that increasing model size yet further pushes the model to rely upon spuriously-correlated features where they carry more signal than the intended features. 
While our conclusions---that a bias for simplicity isn't always desirable---are shared with other works \cite{Shah2020, Dagaev2021, Sagawa2020}, our unique contribution is to show that simplicity bias may be harmful \emph{even} in the balanced data regime and absent of obvious shortcuts, due solely to model-perceived complexity differences \emph{between groups}.

\section{Prelude: Data difficulty is model-specific}\label{sec:difficulty}

This work is predicated on a simple, perhaps even obvious idea: that certain facets of a dataset may be more or less difficult to a given model. 
Whether individual data points, classes, or social groups, identifying what will be difficult to a model isn't always possible ahead of time, nor need it align with human intuition.
As an illustration, we begin with a short investigation of difficulty difference between pairs of classes, though from \cref{sec:amplification} onward we focus on social groups.

\subsection{Not all class pairs are equally difficult}

As a warm-up, we show how a chosen model finds separating certain pairs of classes more difficult than other pairs, by training a ResNet-18 \citep{He2016} classifier on coarse-grained CIFAR-100 \citep{Krizhevsky2009}.
To reduce the confounding effects of covariate shift, we draw ten random train/test splits from the full dataset (i.e. drawn from the original train and tests combined so as to maximize data availability).
We train a randomly-initialized model on each random train split, and measure a class pair's difficulty by its binary classification accuracy on the corresponding test split, having masked irrelevant outputs before the softmax layer.

\Cref{fig:difficulty}{a} shows how the model produces test accuracies that vary between class pairs.
For example, this model achieves near perfect test accuracy on \emph{flowers} vs.~\emph{aquatic mammals}, and substantially lower performance on \emph{non-insect invertebrates} vs.~\emph{insects}.
Crucially, this view of difficulty is just that of this specific model. Were we to perform this evaluation on a different model, our results would differ.

\subsection{\emph{Which} class pairs are difficult depends on the model}
To test whether difficulty is model-specific, we repeat the experiment using the following models: an SVM with an RBF kernel; a 3-layer and a 5-layer fully-connected network; LeNet, a simple CNN \citep{Lecun1998}; AlexNet, a more complex CNN \citep{Krizhevsky2012}; ResNet-34 and -50; and a fully-connected layer over pre-trained representations extracted from ResNet-50 trained using SimCLR \citep{Chen2020} on ImageNet-1K \citep{Russakovsky2015}, and those of RegNet-128Gf \citep{Radosavovic2020} trained with SwAV \citep{Caron2020} on 1 billion public images from Instagram \citep{Goyal2022b} (see \cref{sec:app-model-details} for a full description).

\Cref{fig:difficulty}{b} shows how the rank order correlation of pairwise difficulties varies between models, as measured by Kendall's \(\tau\). 
For a specific example, between the 3-layer FC model and LeNet, there is high (but not perfect) \(\tau\), indicating that broadly what LeNet finds difficult so too does the FC network. 
In contrast, between the RBF SVM and a linear layer on RegNet-128Gf representations, there is much lower (though still positive) \(\tau\), indicating that the pairwise ordering does vary considerably. 
This aligns with recent work \citep{Hacohen2020} showing that the difficulty of individual data points is shared across random initializations of the \emph{same model architecture}, but difficulty is only partially consistent \emph{across architectures}.
While the consistently positive correlations in \cref{fig:difficulty}{b} suggest that difficulty does comprise a data-driven, model-independent component, the lower correlations between certain model pairs (e.g., SVM vs.\ RegNet) confirm that there is also a substantial model-specific factor.

\subsection{Data difficulty is not model difficulty}
We test the relationship between a data-only view of difficulty and model-specific difficulty using the partial least squares (PLS) analysis shown in \cref{fig:difficulty}{c}.
PLS attempts to find low-dimensional projections of both the input and output variables such that their covariance is maximized.
We apply PLS to determine how much a \emph{data-only} difficulty measure can explain a \emph{model+data} measure, where the data-only measure is the rank cosine distance between the input data class means, and the model+data measure is the rank test accuracy (see \cref{sec:app-pls}).
If model difficulty was purely a function of data difficulty, we would expect PLS to find a well fitting linear regression model.
Instead, PLS finds a near-zero fit (\(R^2 = 0.058\)).
From the 1D projection, we see that for some classes (in blue, e.g. \emph{carnivores} vs.~\emph{food containers}), increasing inter-class distance tends to increases binary accuracy, though for many class pairs (in red, \emph{outdoor scenes} vs.~\emph{outdoor things}) the opposite is true.

\noindent \textbf{Summary:} \emph{There is no clear difficulty ordering of class pairs that is consistent across all models. What a model finds difficult is not solely a function of the data, indicating a complex relationship between data and model.}

\section{Neural networks prioritize ``easy''}\label{sec:amplification}

\begin{figure*}[t]
    \centering
    \subfloat[Synthetic task]{
        \includegraphics[trim={0 -1cm 0 0},clip,width=0.22\textwidth]{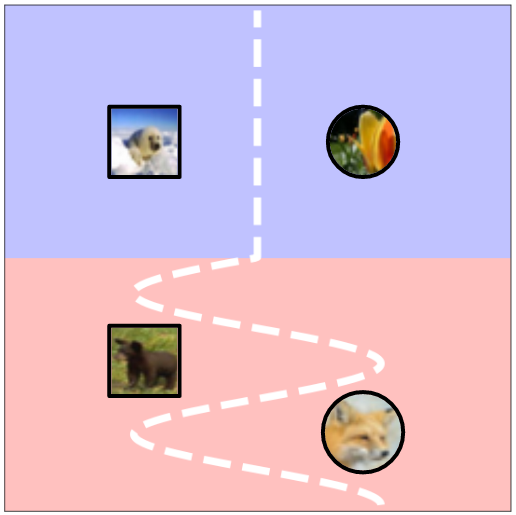}
    }\hfill
    \subfloat[Train accuracy]{
        \includegraphics[trim={0 0 0 0},clip,width=0.24\textwidth]{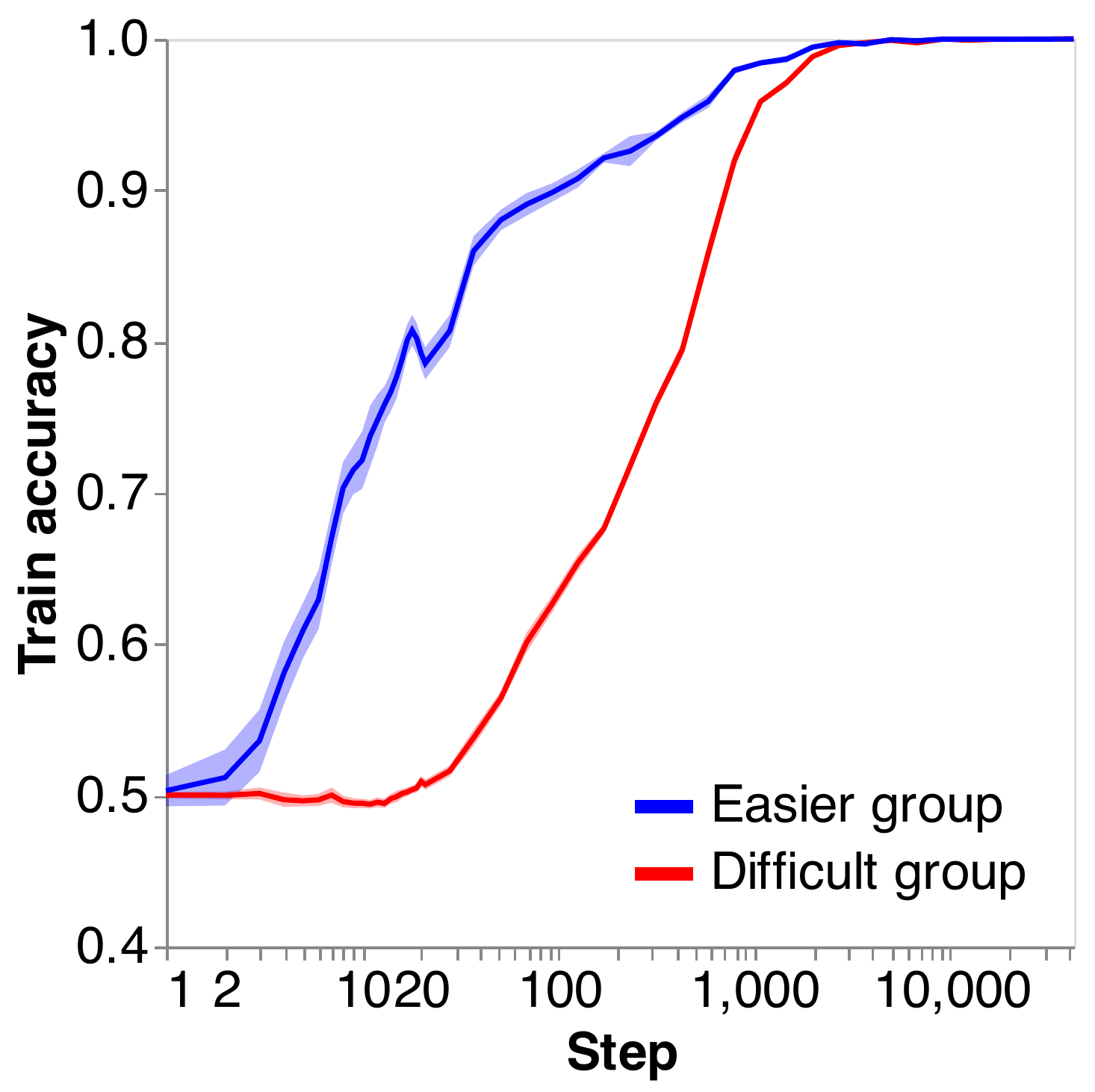}
    }\hfill
    \subfloat[Test accuracy]{
        \includegraphics[trim={0 0 0 0},clip,width=0.24\textwidth]{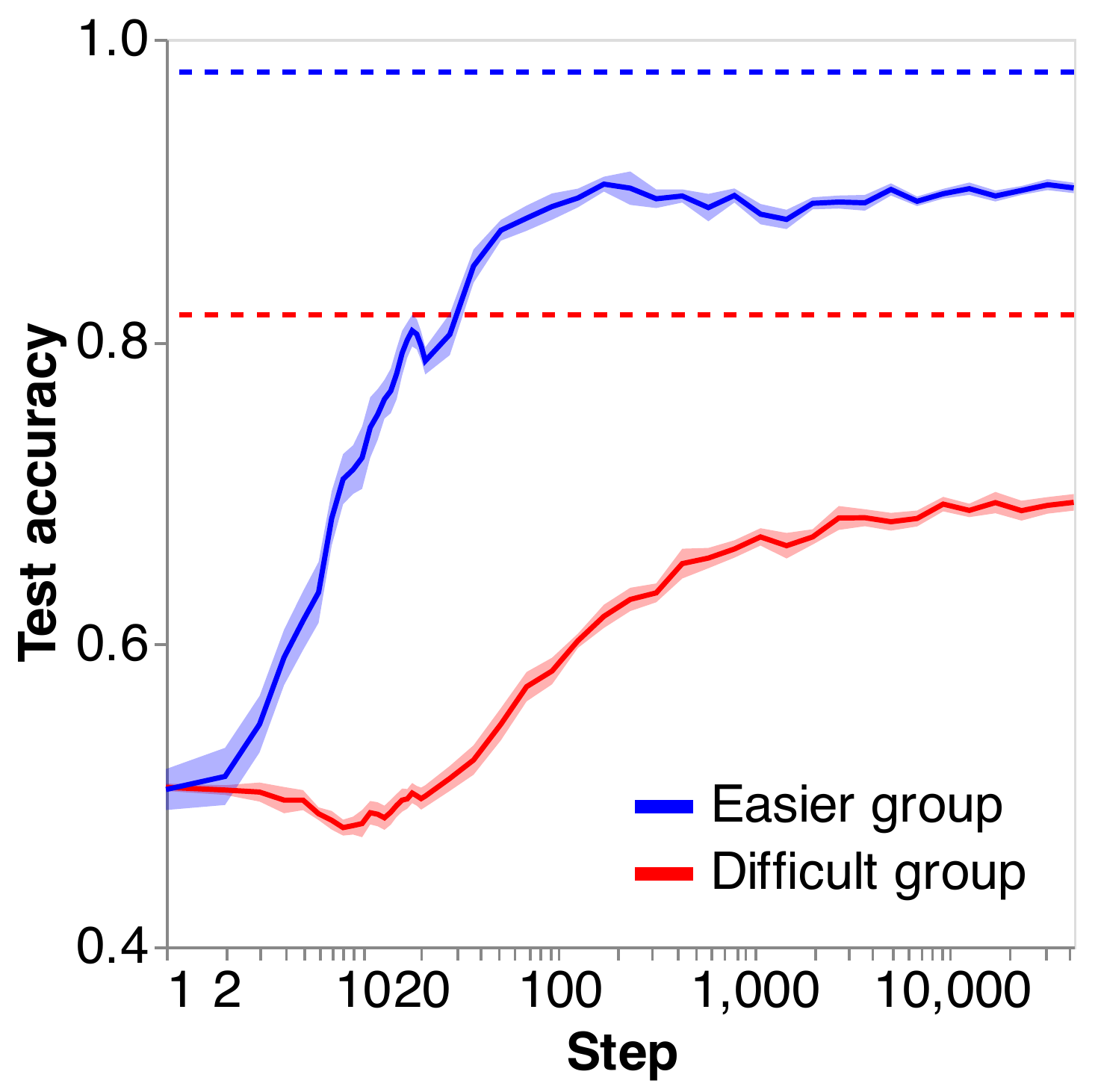}
    }\hfill
    \subfloat[Observed disparity]{
        \includegraphics[trim={0 0 0 0},clip,width=0.24\textwidth]{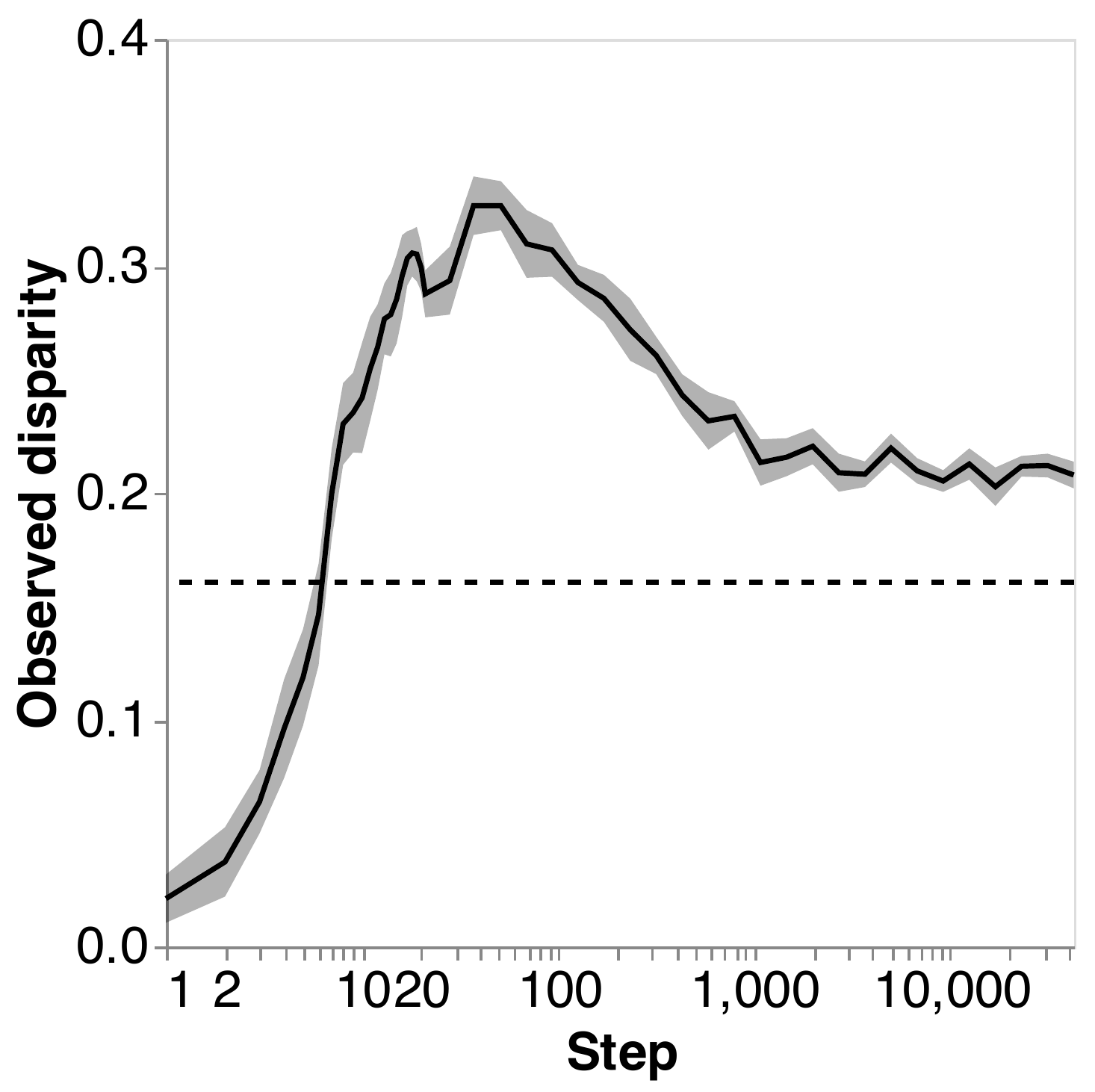}
    }
    \caption{Accuracy and disparity for ResNet-18. \textbf{(a)}~Binary classification (squares vs.~circles) comprising two groups. Easy group (\textcolor{blue}{blue}) is high-accuracy pair \emph{flowers} vs.~\emph{aquatic mammals}. Difficult group (\textcolor{red}{red}) is low-accuracy pair \emph{medium mammals} vs.~\emph{carnivores}. Decision boundary for illustration only; cf.~\cref{fig:sketch-alt}{a}.
    \textbf{(b)}~Train accuracy (solid lines) for complex group is learned slower but both groups reach perfect train accuracy during combined training. \textbf{(c)}~However, test accuracy (solid lines) for complex group is persistently lower during combined training. Dashed lines are test accuracy from single-group training. \textbf{(d)}~Observed disparity \(d\) peaks early in training but large gap persists after training. Black dashed line is estimated disparity \(\tilde{d}\): observed disparity above this line indicates \emph{amplification}. Shading is standard error over 10 runs with different train/test splits.}
    \label{fig:cifar-results}
\end{figure*}

Above we establish how
\textit{difficulty} is a function of both model and data. Now, we turn to identifying which groups neural networks find simpler. To this end, we explore data difficulty as a notion that is relative to a given model. We also quantify how much models selectively prioritize the simple group by way of measurement of the amplification factor.

\subsection{Definitions}
Let \(\mathrm{acc}(X, \mathbf{y}, \mathcal{M})\) be the cross-validated test accuracy on the classification dataset \((X, \mathbf{y})\) of a model \(\mathcal{M}\), and \(\mathcal{M}_{X,\mathbf{y}}\) be a model trained on \((X, \mathbf{y})\). Given two groups \(\alpha\) and \(\beta\), let \((X_{\alpha}, \mathbf{y}_{\alpha})\) and \((X_{\beta}, \mathbf{y}_{\beta})\) denote corresponding  slices of the dataset \emph{each with the same number of samples},\footnote{We enforce equal group sizes to remove the effect of group imbalance. Where possible, groups should also have identical label distributions, though this is not always practical (as is the case in \cref{sec:dollar-street}). In our work, we achieve this via stratified subsampling.} such that a model trained only on group \(\alpha\) is \(\mathcal{M}_{X_{\alpha},\mathbf{y}_{\alpha}}\). 

\textbf{Estimated difficulty disparity}.
First, we define the estimated difficulty disparity \(\tilde{d}\) as the difference in accuracy between a model trained and evaluated on each group in isolation,
\begin{align}
    \tilde{d} = \mathrm{acc}(X_{\alpha}, \mathbf{y}_{\alpha}, \mathcal{M}_{X_{\alpha},\mathbf{y}_{\alpha}}) - \mathrm{acc}(X_{\beta}, \mathbf{y}_{\beta}, \mathcal{M}_{X_{\beta},\mathbf{y}_{\beta}}) \ .
\end{align}

\textbf{Observed difficulty disparity}.
Second, the observed difficulty disparity is the difference in accuracy between groups on a model trained on both groups,
\begin{align}
    d = \mathrm{acc}(X_{\alpha}, \mathbf{y}_{\alpha}, \mathcal{M}_{X,\mathbf{y}}) - \mathrm{acc}(X_{\beta}, \mathbf{y}_{\beta}, \mathcal{M}_{X,\mathbf{y}}) \ .
\end{align}

\textbf{Difficulty amplification}.
We hypothesize that due to simplicity bias, certain models, when given a choice between groups (i.e., during training on both groups), will prioritize the group they find simple, resulting in worse-than-expected performance for the group they find complex (i.e., according to training on each group in isolation).
As such, if the model trained on both groups exhibits worse disparity than when trained in isolation, \(d > \tilde{d}\), we say that the model exhibits difficulty amplification.
Over many model runs, groups, or samples from the dataset we can define an \textbf{amplification factor} \(k = d/\tilde{d}\).

In practice, calculating amplification is a two-stage process.
First, we train \(N\) randomly-initialized models on each group in isolation and compute the average cross-validated test accuracy.
Between each pair of groups, we calculate the estimated difficulty disparity \(\tilde{d}\).
Second, we train a new set of \(N\) models on the full dataset and compute average test accuracy broken out by group. 
For each group pair, we calculate the observed difficulty disparity \(d\), from which we calculate amplification factor \(k\).

\subsection{Simulating difficulty disparity with CIFAR-100}

To test for simplicity bias and measure difficulty disparity in a controlled setting, we design a task based on CIFAR-100 that is group-balanced and absent correlations between group labels and target labels.
We extract the binary test accuracies for each pair of coarse classes and treat their pairwise differences as estimated difficulty disparity \(\tilde{d}\).
We let group \(\alpha\) be the class pair \(y_{0}^{\alpha}, y_{1}^{\alpha}\) with the highest accuracy and \(\beta\) be the lowest \(y_{0}^{\beta}, y_{1}^{\beta}\).
To simulate a binary classification task with two differently-difficult groups, we stitch these pairs together into a single binary task, where \(y_0 = \{y_{0}^{\alpha}, y_{0}^{\beta}\}, y_1 = \{y_{1}^{\alpha}, y_{1}^{\beta}\}\) (see \cref{fig:cifar-results}{a}).
Finally, we train \(N\) models on this task and calculate observed difficulty disparity \(d\).

\subsection{Results}
In \cref{fig:cifar-results}{b} we see the training accuracy of the simple group \(\alpha\) improves much more rapidly than the complex \(\beta\), though both groups reach perfect train accuracy eventually.
In contrast, the test accuracies in \cref{fig:cifar-results}{c} (solid lines) remain notably different at convergence, with the model displaying lower accuracy on the complex group.
This gap is the observed accuracy disparity shown in \Cref{fig:cifar-results}{d}.
Here, we observe that observed disparity peaks after just a few steps, before a slight decline to a plateau.
Supporting our hypothesis, observed disparity remains higher than estimated disparity we would have expected from separate training (the black dashed line).
Replications of this experiment on both Fashion MNIST \citep{Xiao2017} and EMNIST Letters \citep{Cohen2017} show similar results (see \cref{fig:fashion-mnist-results} and \cref{fig:emnist-results}). 

Note that because estimated difficulty disparity is calculated using individual groups, and observed difficulty disparity with all groups combined, the size of the training set differs between these two measures.
Interestingly, \cref{fig:cifar-results}{c} shows higher average accuracy for the single-group training compared to the combined-group training, suggesting that overall dataset size is not responsible for the increased disparity.

\noindent \textbf{Summary:} \emph{This simple experiment shows that models trained across groups with different difficulty do prioritize the simpler group, leading to an outsized observed disparity, primarily driven by under-performance on the more difficult group.}

\section{Amplification factor varies across models}\label{sec:design-decision}

\begin{figure*}[h!]
    \centering
    \subfloat[Observed vs.\ estimated]{
        \includegraphics[trim={0 0 0 0},clip,width=0.24\textwidth]{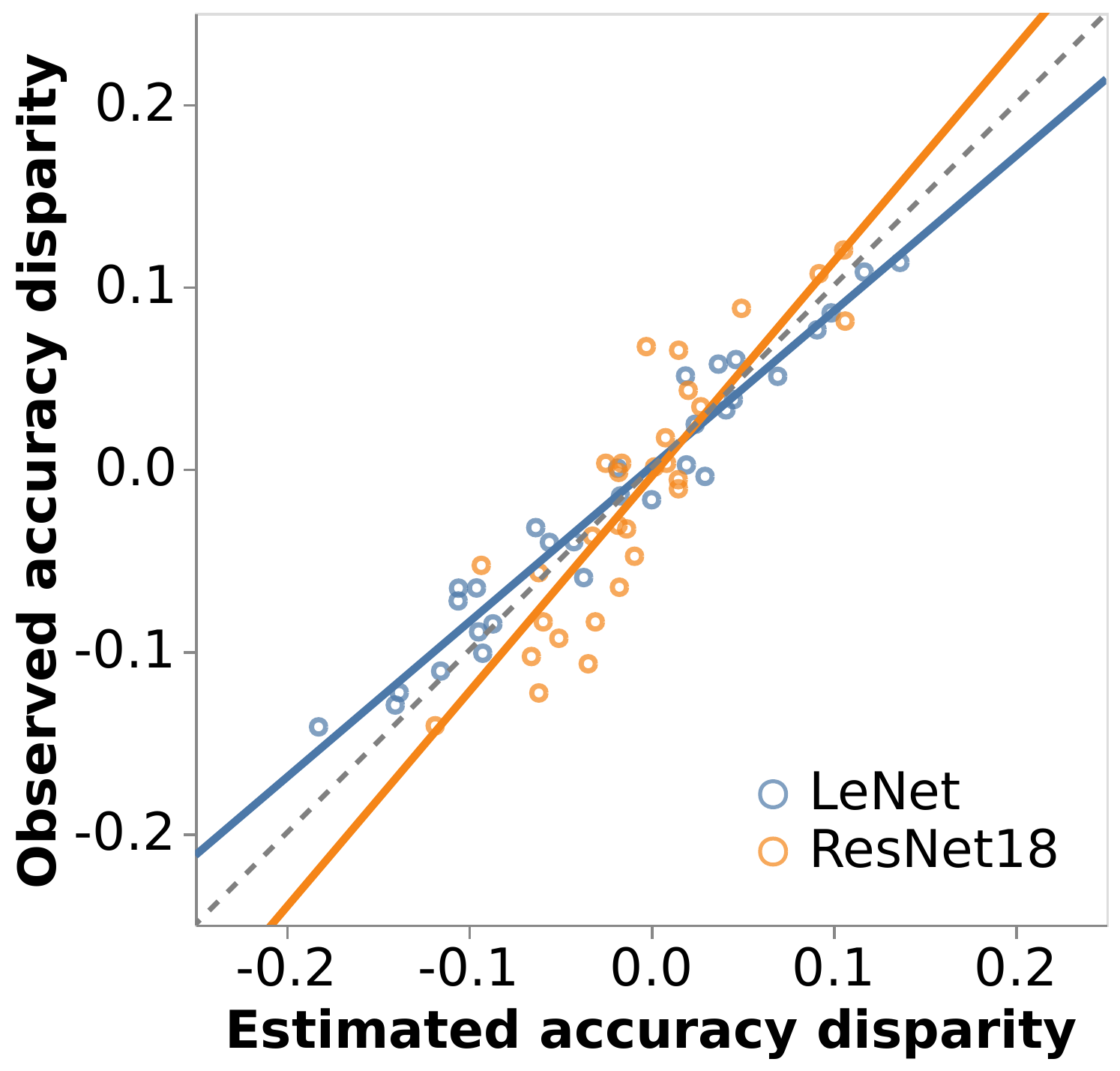}
    }\hspace{20pt}
    \subfloat[Amplification factor]{
        \raisebox{0.45em}{\includegraphics[trim={0 0 0 0.2cm},clip,width=0.24\textwidth]{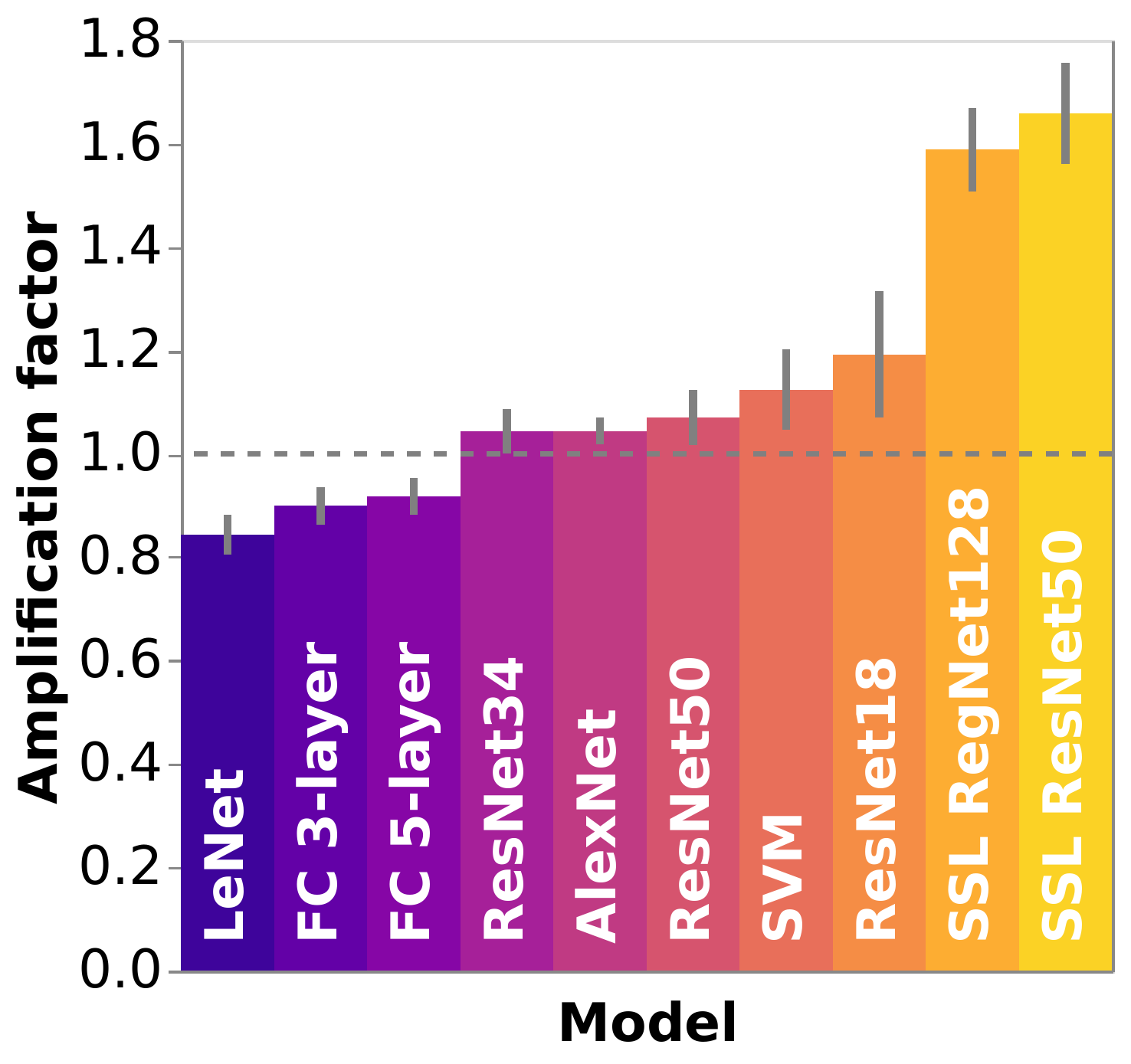}}
    }
    \caption{Observed difficulty disparity as a function of estimated difficulty disparity.
        \textbf{(a)}~\textbf{ResNet-18 amplifies difficulty disparity but LeNet attenuates.} Points are sampled tasks with varying estimated difficulty disparities. Solid line is linear regression. Dashed grey line is amplification factor \(k=1\). \textbf{(b)}~\textbf{Model architecture impacts difficulty amplification factor.} See \cref{fig:app-amplification-r2} for R\(^2\) and \cref{fig:app-amplification-by-test-acc} for raw test accuracies.
    }
    \label{fig:amplification}
\end{figure*}

To quantify the effect of the simplicity bias, we compute an amplification factor by repeating the above experiment using different pairs of classes with different estimated disparity, and compute their observed disparity after combined training.
We retrain on 30 sampled pairs of label pairs, recomputing both estimated and observed difficulty disparity, and apply OLS linear regression to estimate the \emph{amplification factor} (see \cref{sec:app-amplification-factor} for details).
This method can easily be applied to any dataset annotated with group information, by replacing the sampling of pairs of classes with the sampling of different group combinations.
We compute \(k\) for each model listed in \Cref{sec:difficulty}.

Furthermore, we evaluate the effect of model scale on amplification factor by varying the width of ResNet-18; evaluate various settings of weight decay; and evaluate the role of early stopping by computing amplification through training.
Our choice to investigate these three parameters is motivated by their expected effect on simplicity bias.
Following \cite{Kalimeris2019} we expect models to exhibit a stronger preference for simplicity earlier in training, which would often materialize when using early stopping.
Weight decay is a common regularization technique intended to limit overfitting by penalizing excessively complex functions, and the role of width in over-reliance on spurious correlations is reported by \cite{Sagawa2020}.

For a complementary test of the bias \emph{against complexity}, we also try to push the model to choose a more complex solution.
We enforce a Lipschitz constraint by applying a penalty on the norm of the gradients, a technique commonly used to stabilize discriminator training in GANs \citep{Gulrajani2017}.
We add the following penalty term to our loss function \(L\),
\begin{equation}
    L' = L + \lambda ( ||\nabla_{\mathbf{x}} f(\mathbf{x})||_{2} - C )^{2} \ ,
\end{equation}
where \(\nabla_{\mathbf{x}} f(\mathbf{x})\) is the gradient of the network's outputs with respect to its inputs, the penalty coefficient \(\lambda=10\), and \(C\) determines the Lipschitz constraint: a low \(C\) pushes the model towards simpler functions, and high \(C\) towards more complex.

\subsection{Results}

\textbf{Model architecture.} In \cref{fig:amplification}{a}, we illustrate the difference in amplification factor between two models, LeNet and ResNet-18.
We find that the ResNet-18 amplifies disparity by a factor of \(k=1.19\pm0.12\).
In contrast, LeNet diminishes disparity (\(k=0.84\pm0.04\)), resulting in an observed disparity lower than expected.
Thus, from this simple example we show that model choice influences difficulty amplification.
Across the full suite of models (\cref{fig:amplification}{b}) we again see significant variation in amplification factors across the different models, with the certain models attenuating and others amplifying.
\emph{However}, the simpler models all exhibit poor test accuracy averaged over the entire dataset (see \cref{fig:app-amplification-by-test-acc}), offering a candidate explanation for the lack of amplification. 
These models may be too simple to learn the dataset at all, resulting in equally poor performance across all groups. 

\begin{figure*}[h]
    \centering
    \subfloat[Width]{
        \includegraphics[trim={0 0 0 0},clip,width=0.235\textwidth]{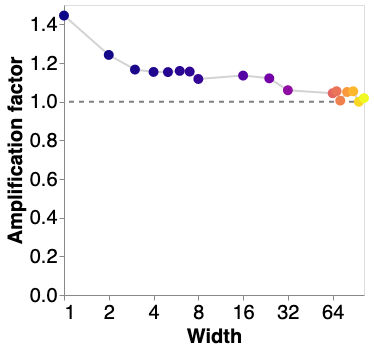}
    }
    \subfloat[Step]{
        \includegraphics[trim={0 0 0 0},clip,width=0.24\textwidth]{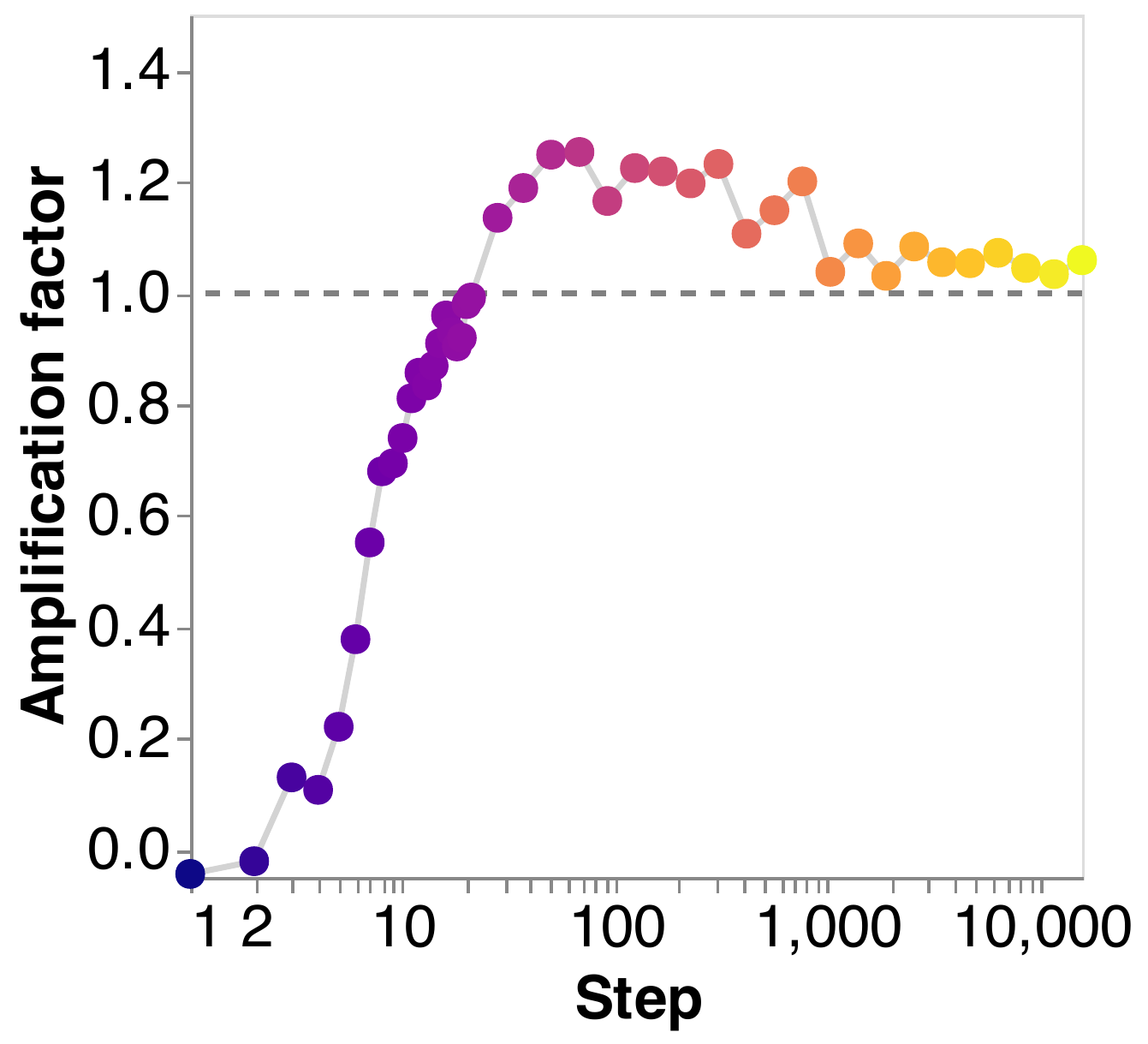}
    }
    \subfloat[Weight decay]{
        \includegraphics[trim={0 0 0 0},clip,width=0.235\textwidth]{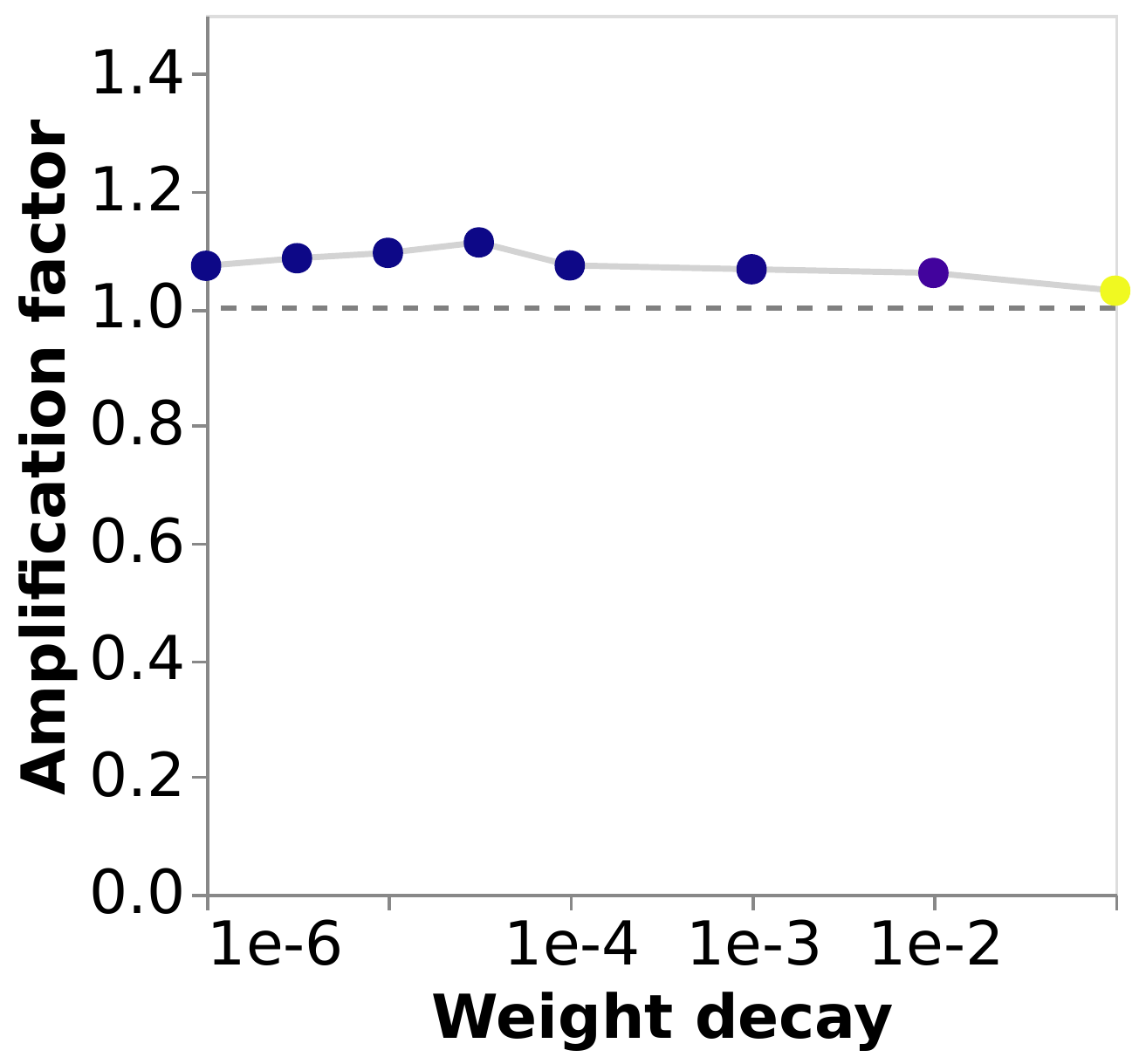}
    }
    \subfloat[Gradient penalty]{
        \includegraphics[trim={0 0 0 0},clip,width=0.235\textwidth]{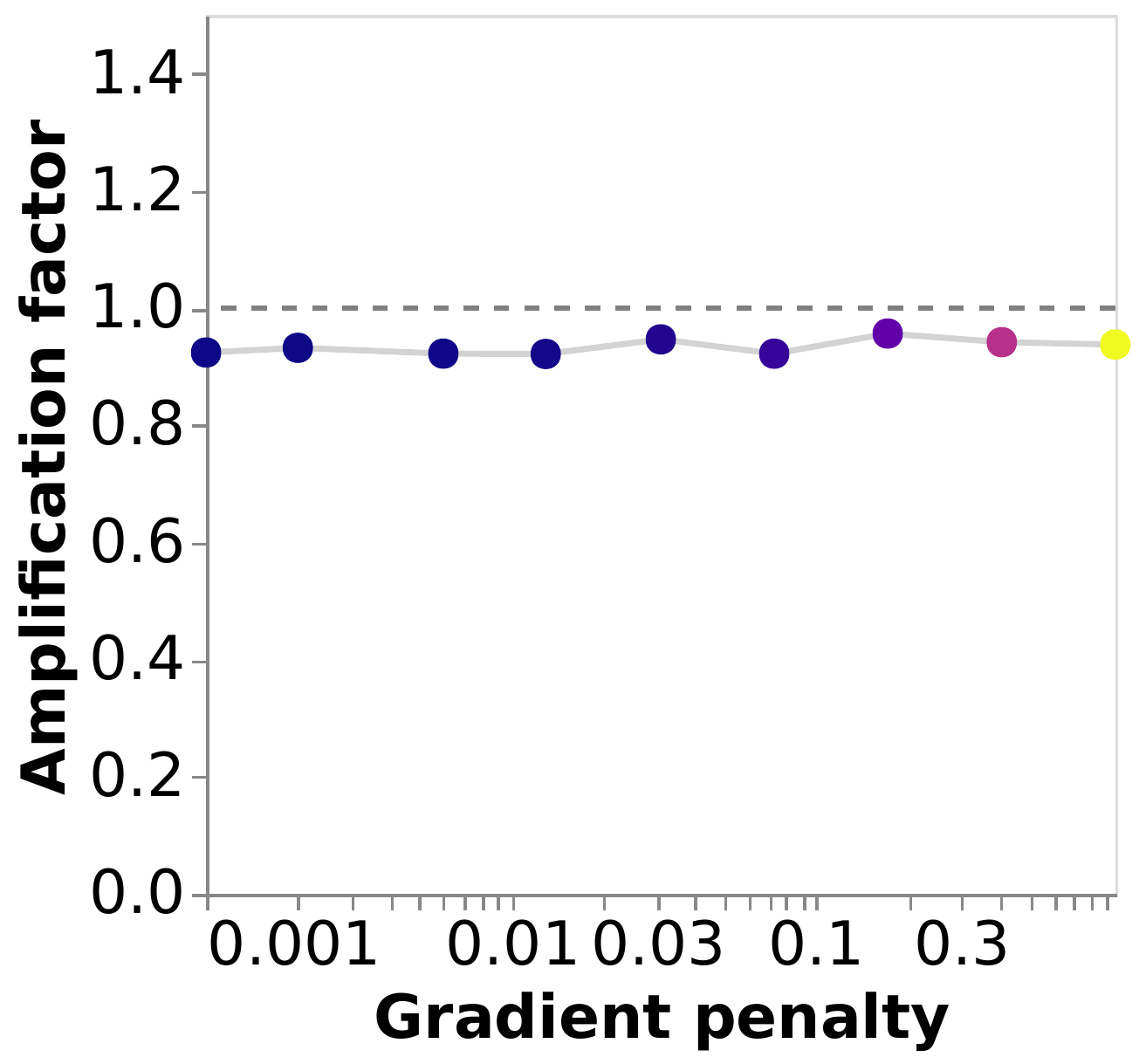}
    }
    \caption{Design decisions influence amplification factor. All panels are a ResNet-18 on 30 sampled synthetic tasks based on CIFAR-100 (see \cref{fig:cifar-results}{a}). \textbf{(a)}~Network width has a negative effect on amplification, reducing it to near 1. \textbf{(b)}~\(k\) peaks early in training before plateauing just above 1. \textbf{(c)}~Weight decay has no effect on \(k\). \textbf{(d)}~Applying a gradient penalty to bias the model toward a \(C\)-Lipschitz functions lowers \(k\). See \cref{fig:app-design-decisions-r2} for \(R^2\).
    }
    \label{fig:design-decisions}
\end{figure*}

\textbf{Width.} However, within a specific architecture, increasing width seems to \emph{reduce} amplification.
\Cref{fig:design-decisions}{a} shows the amplification factor for ResNet-18 rapidly decreasing to almost 1 (no amplification) as network width increases.
These results align with those of \citeauthor{Sagawa2020}, who report that while overparameterization \emph{typically} increases reliance on spurious correlations and increases worst-group error, this effect is \emph{reversed} as groups become more balanced, such that increasing parameter count becomes helpful \citep[e.g.\ fig.~6]{Sagawa2020}.

\textbf{Early stopping.} As training proceeds (\cref{fig:design-decisions}{b}), \(k\) increases to a peak around \(1.2\) early in training, before decreasing to a plateau a little over \(1\). 
This highlights the important role of early stopping in amplifying disparity, particularly in light of prior work arguing that models learn more simple functions earlier in training \citep{Kalimeris2019}. 

\textbf{Weight decay.} \Cref{fig:design-decisions}{c} shows next to no effect of scaling the weight decay parameter.
This is a surprising negative result, as our expectation was that applying stronger weight decay would further bias the model towards the simpler group, increasing amplification.
One possible explanation is the sensitivity of the \(\ell_{2}\) penalty to choices of model and dataset, as reported by \citeauthor{Sagawa2019} \cite{Sagawa2019}.
While the purpose of our work is to introduce the notion of difficulty disparity and difficulty amplification, further research is needed to confirm the role of weight decay across various settings, and its interaction with other implicit regularization schemes.

\textbf{Gradient penalty.} In contrast, applying a penalty to the norm of the gradients, rather than the parameters, is sufficient to lower \(k\) to below 1 for all values of \(C\) considered here.
This suggests that applying a gradient penalty to balance out the implicit bias towards simplicity may be a helpful strategy in combating difficulty disparity.

\noindent \textbf{Summary:} \emph{High-performing models---those optimized for average test accuracy---consistently display difficulty amplification. This phenomenon is exacerbated by early stopping, but may be reduced using a gradient penalty.}

\section{Difficulty amplification has real-world impact}\label{sec:real-world}

To demonstrate the impact of simplicity bias via difficulty amplification, we now present two case studies where observed disparity varies by group, and where observed disparity often exceeds estimated disparity.

\subsection{Age classification on FairFace}\label{sec:fairface}

FairFace \citep{Karkkainen2021} is a dataset of human face images intended for fairness research and audit purposes. 
It comprises a subset of the YFCC-100M Flickr dataset \citep{Thomee2016}, with each sample annotated with perceived age, race, and gender labels, and aims to capture a reasonably balanced distribution with respect to race and gender.
We provide an extended commentary on the nature of the annotations in \cref{sec:discussion-fairface-annotations}, though for our purposes, FairFace serves as a useful illustration of the presence of algorithmic bias even when using a balanced dataset.
Having discarded information on perceived gender, we construct an age classification task, and evaluate performance disparities between groups\footnote{While many datasets that include group labels fail to capture the complex realities regarding why such labels might have been constructed, it remains important to evaluate for whom these models function as intended. See \cref{sec:discussion} for extended commentary on demographic annotations.}, where a group comprises all samples with the same race annotation, which we describe as, for example, ``black-annotated'' or ``white-annotated''.
While FairFace is reasonably balanced, we further apply subsampling in order to precisely equalize the number of samples in each group, and we match the age distribution in each group to remove spurious correlations between race annotation and age annotation.
We train a randomly-initialized ResNet-18 model to classify each sample into one of nine age buckets.
We evaluate models trained on each race-annotation group independently, and models trained on all samples together. 
See \cref{sec:app-fairface} for details.
 
\begin{figure*}[t]
    \centering
    \subfloat[Disparity versus black-annotated group]{
        \includegraphics[trim={0 0 22.5cm 0},clip,height=2.75cm]{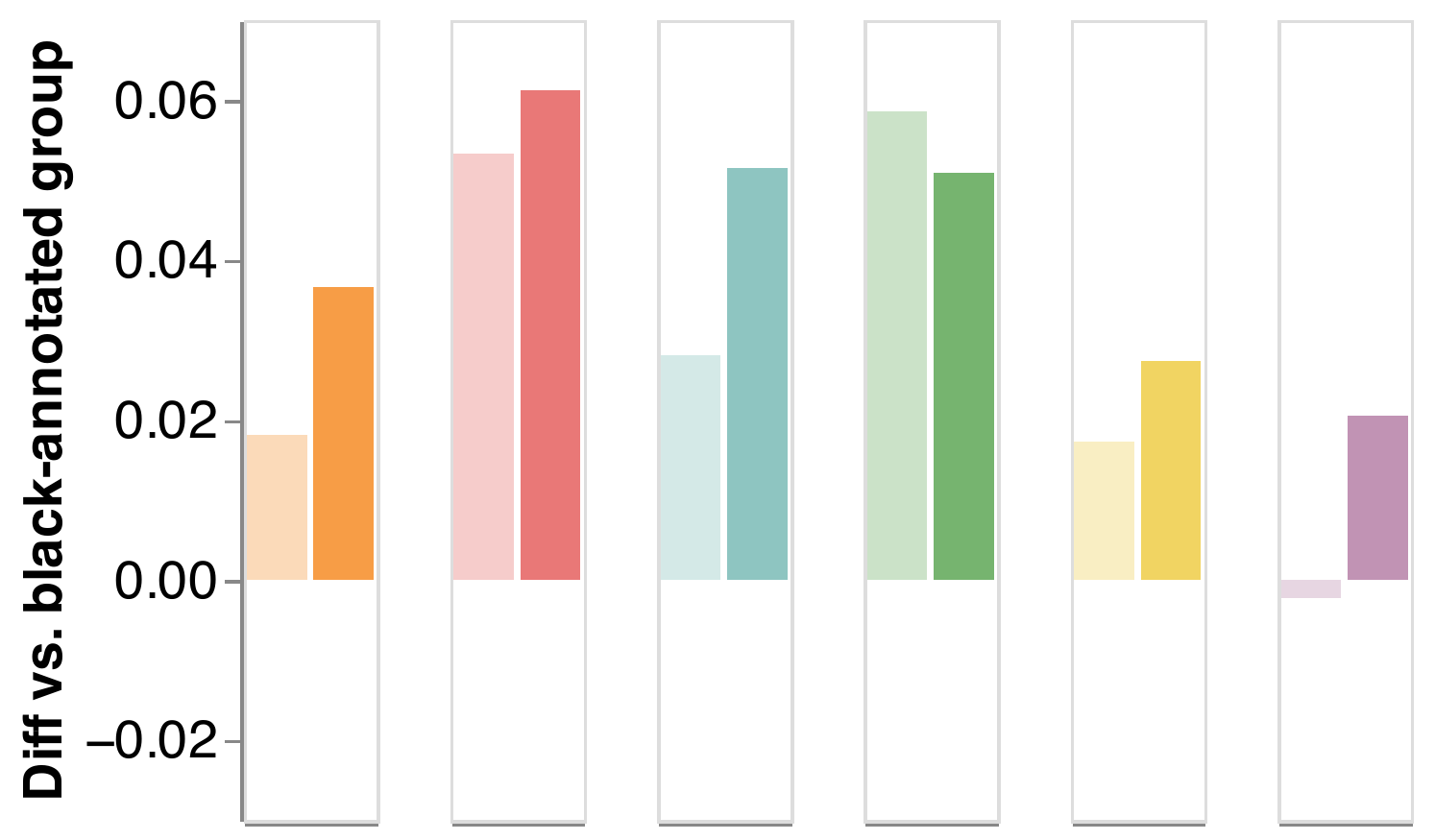}
    }\hspace{20pt}
    \subfloat[Disparity versus white-annotated group]{
        \includegraphics[trim={15.5cm 0 0 0},clip,height=2.75cm]{fig/fairface-age-by-race}
        \hspace*{-2cm}
    }
    \caption{Performance disparity and amplification, measured by difference in mean test accuracy, on FairFace age classification, disaggregated by annotator-perceived race. Translucent bars are estimated disparity from single-group training; opaque bars are observed disparity when trained on entire (balanced) dataset. \textbf{(a)} Performance disparity relative to the black-annotated group. Model demonstrates higher test accuracy for most other race-annotation groups (opaque). Compared with single-group training (translucent), this particular model consistently amplifies the disparity, resulting in poorer-than-expected model performance for the black-annotated samples. \textbf{(b)} Performance disparity relative to the white-annotated group. In contrast, for white-annotated samples, observed disparity (opaque) is consistently lower than estimated disparity (translucent, indicating disparity attenuation.}
    \label{fig:fairface}
\end{figure*}

\Cref{fig:fairface}{a} shows the observed (opaque) and estimated (translucent) performance disparity between black-annotated images and other race-annotation groups. 
For all but one comparison, the observed disparity exceeds the estimated disparity, indicating the presence of difficulty amplification. 
Due to our balanced and distribution-matched dataset construction, we can confidently say that this is not a result of an imbalanced dataset or group/label associations.
In practical terms, this figure shows that this particular model, ResNet-18 trained from scratch, performs worst on black-annotated samples, and crucially that this performance gap is worse-than-expected: the model has selectively prioritized other groups that it finds simpler.

In contrast, \Cref{fig:fairface}{b} shows the same comparison of each race-annotation group, but compared against a white-annotated baseline instead.
Here, we observe the opposite effect compared with \cref{fig:fairface}{a}. 
Observed disparity (opaque) is always lower than estimated disparity (translucent), indicating attenuation of difficulty for this group relative to the given model. 
Unlike the worse-than-expected disparity the model exhibits on black-annotated samples, for white-annotated samples the model demonstrates lower-than-expected disparity.
These results reinforce that performance disparities are complex and hard to predict, and may have heterogeneous impact across different groups.

\subsection{Object classification on DollarStreet}\label{sec:dollar-street}

Dollar Street \citep{DollarStreet} is a dataset of geographically-diverse images spanning a broad range of household incomes.
We use the labels associated with each image in a 138-class object classification task, where group information is household income quartile.
We explicitly rebalance the dataset via subsampling to ensure each group has the same number of data points, though in this experiment, we don't match the label distributions between groups due to limited data availability for certain group \(\times\) label pairs.
We evaluate models trained on each income quartile independently, and models trained on all quartiles together. 
We train single-layer FC networks on representations extracted from ResNet-18 pretrained on ImageNet-1K.
See \cref{sec:app-dollar-street} for details, including discussion on the potentially confounding effects of pretraining. 

\begin{figure*}[t]
    \centering
    \subfloat[Disparity vs. 1st quartile]{
        \includegraphics[trim={0 0 15.5cm 0},clip,height=2.75cm]{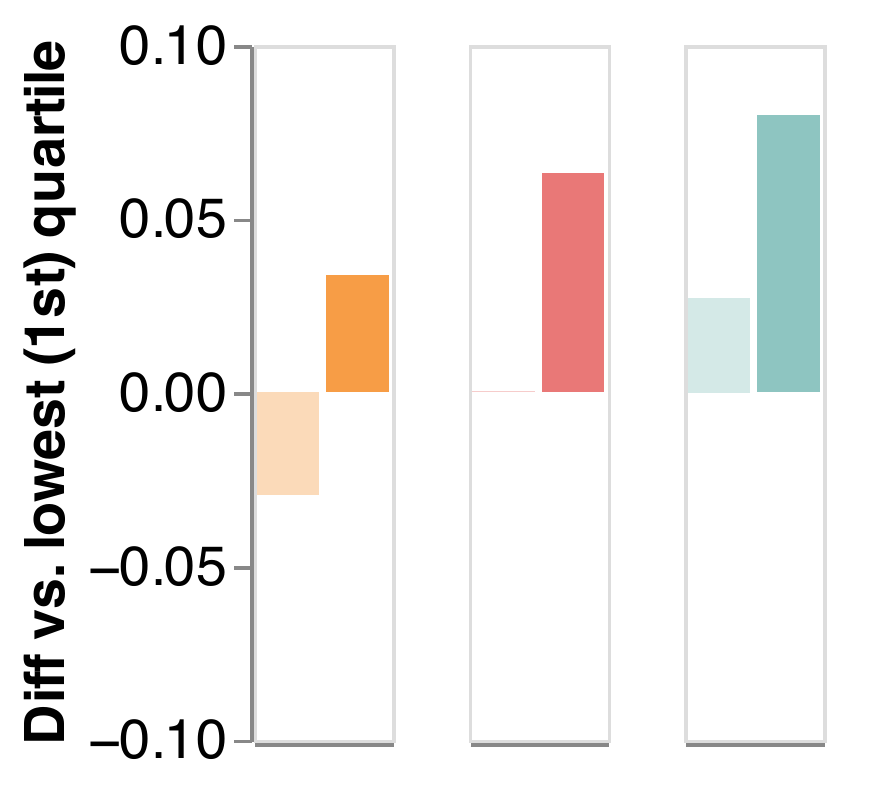}
    }\hspace{20pt}
    \subfloat[Disparity vs. 4th quartile]{
        \includegraphics[trim={9cm 0 0 0},clip,height=2.75cm]{fig/dollar-street-by-income}
        \hspace*{-2cm}
    }
    \caption{Performance disparity and amplification, measured by difference in mean test accuracy, on Dollar Street object classification, disaggregated by household income quartile. Translucent bars are estimated disparity from single-group training; opaque bars are observed disparity when trained on the entire (balanced) dataset. \textbf{(a)} Performance disparity relative to lowest (1st) income quartile. Model demonstrates higher test accuracy for most other income quartiles (opaque). Compared with single-group training (translucent), this model consistently amplifies the disparity, resulting in poorer-than-expected model performance for the lowest income quartile. \textbf{(b)} Performance disparity relative to highest (4th) income quartile. In contrast, for the highest income group, observed disparity (opaque) versus the 1st quartile is larger than estimated (translucent), whereas vs. the 2nd and 3rd quartiles the performance disparity is slightly attenuated.}
    \label{fig:dollar-street}
\end{figure*}

\Cref{fig:dollar-street}{a} shows the observed (opaque) and estimated (translucent) performance disparity between images from households in the lowest (1st) income quartile, compared against other quartiles. 
As with FairFace, observed disparity consistently exceeds the estimated disparity, indicating difficulty amplification. 
In contrast, \cref{fig:fairface}{b} compares each quartile against the highest (4th) income quartile, and presents a mixed picture. 
Here, the observed performance gap between 1st and 4th is amplified (though to the benefit of the 4th quartile, and the detriment of the 1st), whereas the observed disparity between the 4th and the two middle quartiles is slightly attenuated. 
These results again confirm that performance disparities persist in the balanced data setting, and performance disparities may be selectively amplified depending on one's choice of model. 
In the case of this Dollar Street example, this selective amplification results in worse-than-expected performance disparities and excessively degraded performance on images from the lowest-income households.

\noindent \textbf{Summary:} \emph{Across two tasks on two different datasets, models exhibit selective difficulty amplification, resulting in worse-than-expected performance disparity for certain groups. On Dollar Street, this is a real-world impact on income disparity, and on FairFace, this manifests as racial performance disparity.}

\section{Mitigating difficulty amplification}\label{sec:mitigation}

The case studies with Dollar Street and FairFace demonstrate that neither a balanced dataset nor equivalent distributions across groups are sufficient to preclude performance disparities. 
Moreover, they demonstrate the heterogeneous impact these disparities can have, and their unpredictable effects (e.g.\ our ResNet-18 model amplifies performance disparity on black-annotated samples, but not on white-annotated samples). 
What should one do when faced with such a scenario?
Having audited our dataset, having striven for balance, imagine we find that our chosen model finds one group harder than the others.
Here, we explore two potential remedies. 
First, we try additional data collection for the group the model performs poorly upon.
Second, in the event this is not possible, we consider oversampling the challenging group such that challenging examples a more likely to be included in a given mini-batch.

We evaluate both mitigation strategies on the same FairFace setup as described in \cref{sec:fairface}. 
As a result of our explicit re-balancing via subsampling, we conveniently have access to previously-discarded samples for various groups. 
To counter poor model performance on the black-annotated group, we add the full set of black-annotated samples.
This results in an \emph{imbalanced} dataset, where all other groups are balanced but black-annotated samples are overrepresented by a factor of approximately 1.6.
For our oversampling experiment, we imagine a setting where further data collection is not possible, and so return to the balanced data setting. 
Instead, we assign twice the weight to each black-annotated sample, such that a member of this group is twice as likely to be included in a mini-batch than any other group.

\begin{figure*}[t]
    \centering
    \subfloat[Obs.~disparity with additional data]{
        \includegraphics[trim={0 0 0 0},clip,height=2.75cm]{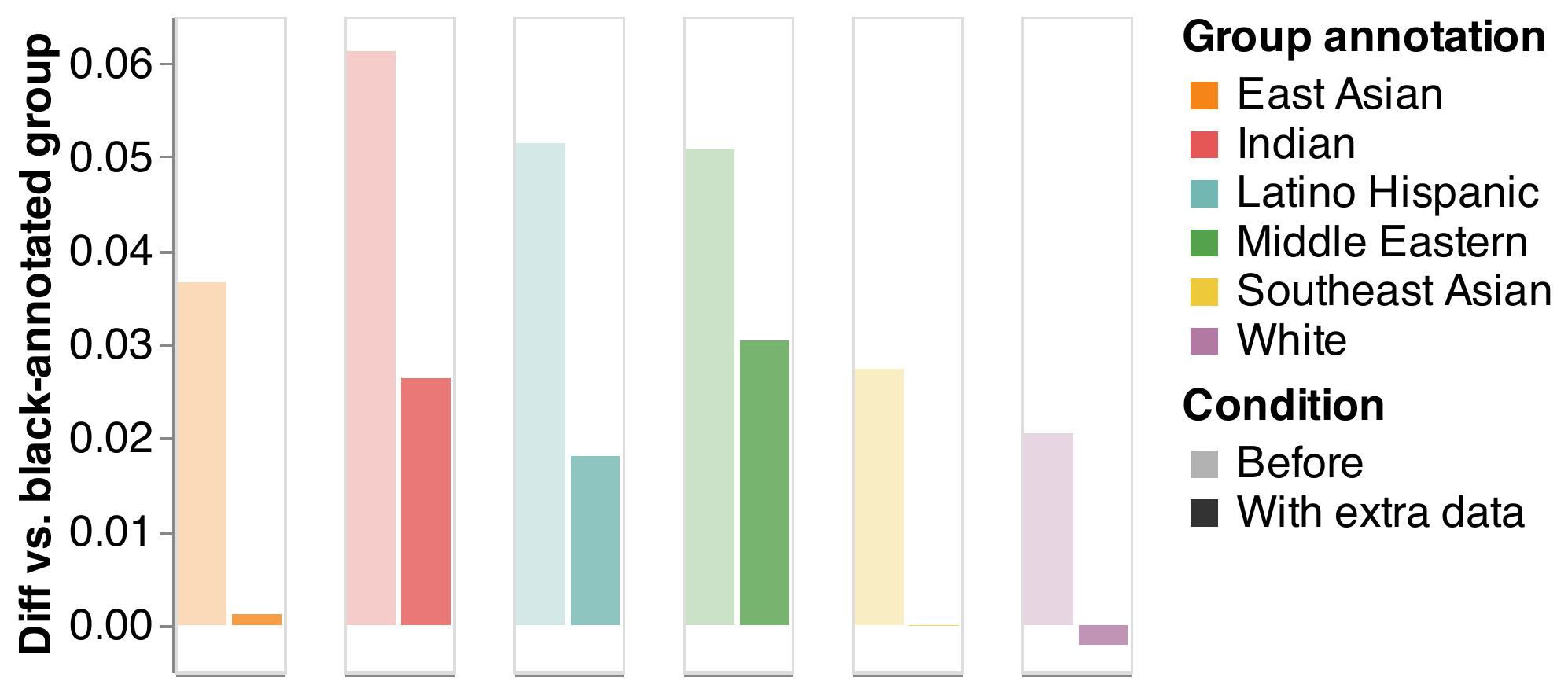}
    }
    \hspace{20pt}
    \subfloat[Obs.~disparity with oversampling]{
        \includegraphics[trim={0 0 0 0},clip,height=2.75cm]{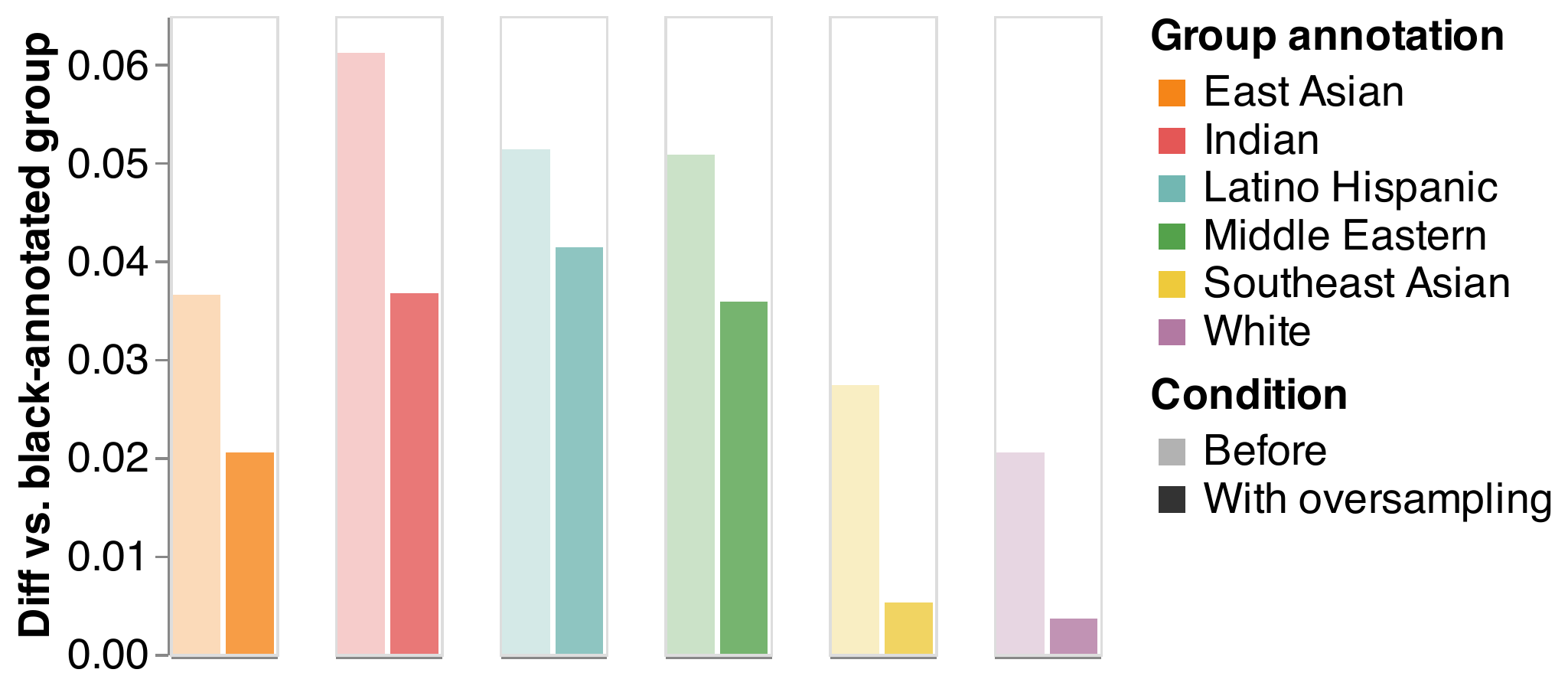}
    }
    \caption{Two mitigation strategies for performance disparity. \textbf{(a) Collecting additional data}. After increasing the amount of data available for the black-annotated group by a factor of 1.6, we find observed disparity is substantially reduced.
    \textbf{(b) Oversampling}. Where additional data collection may not be possible, oversampling the black-annotated group by a factor of 2 also reduces observed disparity, though to a lesser extent than data collection.
    Translucent bars are observed disparity without intervention (also shown in \cref{fig:fairface}{a}); opaque bars are with mitigation strategy applied.}
    \label{fig:fairface-mitigation}
\end{figure*}

The results of these experiments can be seen in \cref{fig:fairface-mitigation}.
\Cref{fig:fairface-mitigation}{a} shows a clear effect of additional data collection, reducing the observed disparity (opaque bars) versus the balanced baseline (translucent) relative to all other groups.
Thus, we conclude that where performance disparity is present even with a balanced dataset, a helpful mitigation strategy might be to extend data collection for those groups suffering worse performance.
Where previous work has suggested collecting additional data to achieve a balanced dataset (e.g. \cite{Dutta2020}), here our results suggest we take one step further and construct explicitly unbalanced datasets, though skewed in favor of the groups our chosen model finds challenging.
Such model-dependent data collection entails a feedback cycle of model selection, model evaluation, and targeted data collection that is uncommon in contemporary machine learning practice, and presents a challenge to the use of standardized, off-the-shelf training sets.

In \cref{fig:fairface-mitigation}{b} we show a clear effect of oversampling in reducing observed disparity (opaque bars) versus the uniformly sampled baseline (translucent), though a smaller effect than that of additional data collection.
These results indicate that where additional data collection may be undesirable, oversampling the groups the model finds challenging may be a viable alternative.
We note that oversampling a minority class is a common fairness strategy when handling performance disparities caused by under-representation in a dataset \citep{Idrissi2022, Wang2020}.
However, in our experiments our signal for oversampling is not relative group size, but performance gap.
Starting from a balanced dataset, we increase the sampling likelikhood of samples in the black-annotated group because of the observed performance disparity, not because of a difference in group size.
As with additional data collection, these results highlight the necessity of a tight and iterative loop involving model development and fairness evaluation.

\noindent \textbf{Summary:} \emph{Faced with observed performance disparity that persists with balanced data, an effective mitigation strategy may be to collect additional data for the groups the model performs poorly upon. Should this not be possible, oversampling may be a suitable alternative.}

\section{Discussion}\label{sec:discussion}

\subsection{Auditing for bias}
At its heart, our work presents yet another way in which models exhibit bias and performance disparities across demographics.
A frequent refrain in the ML community is that such disparities are the fault of the data, rather than algorithmic bias \citep{Hooker2021}. 
Indeed, a series of thorough audits have revealed that popular datasets under-represent minoritized groups \citep{Shankar2017, Stock2018, Buolamwini2018, deVries2019, Dulhanty2019, Wilson2019}; reify harmful associations and perpetuate stereotypes \citep{Bolukbasi2016, vanMiltenburg2016, Garg2018, Dixon2018, Birhane2021, Raji2021b}; and operationalize concepts such as gender and race in a way that applies a veneer of ``objectivity'' to socially-constructed and culturally specific concepts \citep{Keyes2018, Paullada2021, Denton2021, Raji2021a}.
Fixing these issues at the level of the data may not even be possible, for example it is often undesirable to collect the demographic information needed to ensure balance in the first place \citep{Veale2017, McKane2021, Hooker2021}.
That being said, acknowledging issues with our use of data does not absolve all that comes after, as exemplified by bias amplification \citep{Zhao2017, Wang2021, Hall2022}.
Here, in support of the role of \emph{post-training} audit, we choose the setting where the data is ``perfect'', in that it is both explicitly balanced, and groups and labels are decorrelated.
The variability of both difficulty disparity and amplification from model to model is a strong reminder that both those who develop and deploy ML systems must take action to ensure their fairness.

\subsection{Measuring difficulty}\label{sec:app-measuring-difficulty} 
In this work we \emph{choose} to measure difficulty using cross-validated test accuracy, averaged over all samples in a group or class.
While it may be possible to rewrite the specific results above in terms of accuracy disparity, we instead refer to difficulty disparity because our core claims involve relative, model-perceived group difficulties, and we expect difficulty amplification to also occur in settings where accuracy is not an appropriate performance metric.

Recent works have investigated alternative methods for quantifying model-specific example difficulty, including loss \citep{Arazo2019, Han2018} and prediction disagreement between models \citep{Simsek2022}, mini-batches \citep{Chang2017}, and throughout training \citep{Toneva2018, Swayamdipta2020}.
\cite{Hooker2021} identifies samples that are often forgotten after compression.
Applying these sample-level measures to evaluating group-level difficulty disparity remains an interesting future direction.

\subsection{Fairness definitions}
By discussing issues of bias and disparity, we engage in a broader discussion about fairness in ML systems.
Here, we follow others in focusing on the performance gap between groups \citep{Dwork2012, Hardt2016, Kale2017, Agarwal2018, Khani2019, Goyal2022a}, though an alternative approach would be to focus explicitly on worst-group performance instead \citep{Mohri2019, Sagawa2019, Zhang2020}.
Others rely upon counterfactual fairness \citep{Kusner2017, Kilbertus2017, Loftus2018}, according to which a ``fair'' system reaches the same decision on two otherwise identical individuals belonging to different protected groups, though this draws increasing criticism due to its requirement that concepts such as race or gender both be well-defined \citep{Benthall2019} and can be changed while only \emph{minimally} impacting other attributes \citep{Hu2020, Hanna2020}.
Our aim in this work is not to use a metric by which to deem systems fair or unfair, but to highlight the possible role of model bias---in this case, due to preference for simplicity---that will have subsequent fairness impacts.
Even assuming a satisfactory yardstick by which to measure, and a model accordingly deemed fair, fairness is of course not necessarily implied.
When situated within a broader societal context, any model can be put to harmful use, and it is a common pitfall of the ML community to narrowly situate our work inside neatly-defined abstractions \citep{Selbst2019}.

\subsection{Spurious correlations}
Similarly, a key ambition of our work is to push research into sources of bias outside of the typical characterizations: spurious correlations and under-representation.
Indeed we suggest that reducing the study of model bias to these two dimensions is an instance of excessive abstraction through formalization \citep{Selbst2019}.
By focusing on the settings where these issues are resolved, we hope that future research can take a more nuanced look at the biased behavior of models where not \emph{obviously} the result of a data issue.
A plausible outcome of this kind of research could be that in certain situations ML might not be appropriate \emph{at all}, if we can't guarantee that the system won't develop unpredictable and hidden biases.

\subsection{FairFace race and gender annotations}\label{sec:discussion-fairface-annotations}
While we select FairFace as a helpful proof-of-concept for the idea of difficulty amplification, there are aspects of the dataset's annotation practices that make for challenging, if not uncomfortable, usage. 
Though we don't make use of so-called gender annotation, we wish to reinforce that gender is not a binary and is not objectively externally-perceivable.
This is not just a matter of ``subjectivity'', a point which \citeauthor{Karkkainen2021} themselves gesture towards, but the more fundamental idea that whether or not the annotators reach consensus, their very act of assessing gender reifies a socially-constructed concept. 
Moreover, the use of binary labels is by design exclusive of non-binary and gender-nonconforming people, groups already overlooked by machine learning research \citep{Keyes2018}.

As for race annotations, again \citeauthor{Karkkainen2021} appear to give little thought to the socially-constructed nature of race.
Appealing to a supposed objectivity, they first take their initial set of race-annotation groups from the U.S. Census Bureau, before going on to further break down certain groups (such as East and South East Asian) because ``they
look clearly distinct'' (p.~1550), and discarding Native Americans, Hawaiians and Pacific Islanders due to limited data availability. 
At the point of annotation, no information is presented as to whether the annotators were trained or presented with reference images, nor is consideration given to the social and cultural contexts of the annotators which would naturally influence their understanding of racial groupings. 
While \citeauthor{Karkkainen2021} present a well-motivated argument for focusing on race rather than the easy-to-compute skin tone used in similar research---due on the one hand to the effect of external lighting conditions and on the other to the high variance of skin tone within racial groups---if one is to focus on race-based performance disparities rather than skin-tone based disparities, significantly more attention must be paid to the partial and culturally-specific way in which race is constructed and deployed.

We have chosen to use FairFace as it presents a \emph{mostly} balanced dataset that allows us to explore the effect of simplicity bias on a dataset of human images, annotated with a fairness group label (in this case, race) that is widely understood as being a source of machine learning disparity.
That said, the above comments regarding data annotation practices should be considered a significant caveat of our real-world case study.

\subsection{Limitations}
Our primary aim is to further highlight the key role of the model in accuracy disparity.
We do however assume access to group information for audit purposes, which may not be available in many realistic scenarios, nor desirable to obtain.
We intentionally choose to explore the balanced dataset setting, though separating difficulty disparity from other sources of bias may be difficult in practice.
Future work may seek to explore a broader array of model families, and a more detailed investigation of the role of different regularization techniques.
Our exploration of the role for additional data collection and oversampling, presented as \emph{candidates} for potential mitigation strategies, are not intended be readily-deployed solutions to alleviate model bias. 
Instead, both potential strategies are presented as evidence in support of the unique nature of difficulty amplification, and on the need for model-specific bias mitigation strategies.

\section{Conclusion}\label{sec:conclusion}

We have argued that what a model finds difficult is not simply a function of the data, but a function of both model and dataset.
This is particularly a problem in a fairness context if difficulty is correlated with group information.
We have found that certain models further amplify difficulty disparity, resulting in observed difficulty disparity over and above estimated difficulty disparity, as a result of the bias of certain models towards easy examples.
Difficulty amplification varies with model architecture, model scale, training time and regularization strategy, and seemingly innocuous design decisions can have a substantial and counter-intuitive impact.
We have shown how difficulty disparity and amplification take place in the Dollar Street setting, where our simple model is biased against images in the low income quartile, and in the FairFace setting, where our exhibit the worst age classification performance on black-annotated samples. 
Through our explorations of two candidate mitigation strategies in this real-world setting, we suggest that explicitly collecting additional data for complex groups, or if not possible, oversampling, may be helpful tools for mitigating performance disparities that persist in the balanced data regime.
Taken together, our results highlight the key role of the model---above and beyond the dataset---in creating group disparities.


\begin{acks}
We are grateful to Randall Balestriero, Diane Bouchacourt, Melissa Hall, Neil Lawrence, David Lopez-Paz, Ari Morcos, Mohammad Pezeshki, Berfin Simsek, Arjun Subramonian, Nicolas Usunier, Adina Williams and Badr Youbi Idrissi for helpful discussions, and to the anonymous FAccT reviewers for their considered and constructive feedback. 
\end{acks}

\newpage

\bibliographystyle{ACM-Reference-Format}
\bibliography{bell-sagun-simplicity.bib}


\newpage

\appendix

\setcounter{figure}{0}

\makeatletter 
\renewcommand{\thefigure}{S\@arabic\c@figure}
\makeatother

\section{PLS Methodology}\label{sec:app-pls}

Given \(m\) classes, we extract \(d=m\frac{m-1}{2}\) binary accuracies corresponding to each possible pairing, and convert them to rank orders.
We repeat this for each of the \(n\) models under consideration, resulting in a \(d \times n\) matrix of difficulty ranks.
We construct a \(d \times 1\) data difficulty matrix from the cosine distance between the mean of each class.

We fit a partial least squares regression model to both the model difficulties and the data difficulties using \emph{scikit-learn} \citep{scikit-learn}.
For visualization purposes we use a single component, though in subsequent tests we find no difference in model fit when increasing the number of components.
We evaluate how well the data difficulty is explained by the model difficulty using \(R^2\). 

\section{Model architecture and hyperparameters}\label{sec:app-model-details}

\subsection{SVM}

For the SVM we use an RBF kernel with hyperparameters \(C=1.0\) and \(\gamma=\frac{1}{3072}\), using the \emph{scikit-learn} implementation.

\subsection{Neural networks}\label{sec:app-model-details-nn}

All models are implemented in \emph{PyTorch} \citep{Paszke2019} with \emph{TorchVision}.
Models are trained to minimize cross-entropy loss using SGD with learning rate \(0.01\), momentum \(0.9\), weight decay of \(0.0001\) for \(500\) epochs with batch size \(128\).

\textbf{FC.} The fully-connected networks are either 3 or 5 hidden layers with 256 units and ReLU activation.
Batch normalization is applied to the inputs. 

\textbf{LeNet.} LeNet \citep{Lecun1998} is a simple CNN, with two convolutional layers interleaved with max pooling, three fully-connected layers, and ReLU activation function.

\textbf{AlexNet.} AlexNet \citep{Krizhevsky2012} is a deeper CNN, with five convolutional layers interleaved with max pooling, three fully-connected layers, and ReLU activation function.

\textbf{ResNet.} We use 18-, 34- and 50-layer variants of the variable-width ResNet \citep{He2016} implementation introduced in \citep{Sagawa2020}.

\textbf{SSL.} For both SSL models, we extract final-layer representations for each data point from an SSL-pretrained model.
We pass these representations through a 1-layer FC network as described above. 
Representations are extracted from one two models.
The first is from a ResNet-50, pretrained with SimCLR \citep{Chen2020} on ImageNet-1K \citep{Russakovsky2015}. 
The second is from a RegNet-128Gf model \citep{Radosavovic2020} trained with SwAV \citep{Caron2020} on 1 billion public images from Instagram \citep{Goyal2022b}. 
Representations were extracted using VISSL \citep{Goyal2021} from models  publicly available in the model zoo.

\section{Dollar Street experiment}\label{sec:app-dollar-street}

\subsection{Dataset}
Dollar Street\footnote{\url{https://www.gapminder.org/dollar-street}} is a dataset of geographically-diverse images spanning a broad range of household incomes.
Dollar Street comprises 23724 RGB \(480\times480\) images of objects and people in everyday environments around the world, each associated with one of 138 class labels.
For our purposes, we discard geographic information and use income quartiles as group label. 

Throughout this work, we have endeavored to remove bias resulting from group imbalance and label/group correlation.
However, in the Dollar Street example we introduce an additional possible source of bias via ImageNet-1K pretraining.
ImageNet-1K significantly under-represents many social groups \citep{Dulhanty2019} and geographies \citep{Shankar2017, deVries2019}, and exhibits harmful associations between race and certain class labels \citep{Stock2018}.
Geographic under-representation is a plausible reason for income-quartile difficulty disparity. This, however, cannot explain difficulty amplification.

\subsection{Model}
We follow the representation extraction method outlined in \cref{sec:app-model-details-nn}, though we use the representations from supervised learning models rather than SSL.
Specifically, we extract from a ResNet-50 trained with supervised learning on ImageNet-1K \citep{Russakovsky2015}.
We train a single layer fully-connected network of varying width, following the standard SGD training regime specified above.

\section{FairFace experiment}\label{sec:app-fairface}

\subsection{Dataset}

FairFace \citep{Karkkainen2021} is a publicly-available dataset of face images for fairness audit and evaluation, intentionally constructed to be approximately racially balanced.
The dataset is a subset of the YFCC-100M Flickr dataset \citep{Thomee2016}, and comprises 97698 RGB \(224\times224\) images of human faces, each annotated with age, race, and gender groups as perceived by Amazon Mechanical Turk annotators, according to majority vote among a committee of three.
Images are labeled with perceived age from 0-2 years, 3-9, 10-19, and in 10 year increments until 70+.
Gender scores are a binary of male or female, and race annotation is a single selection from one of White, Black, Latino Hispanic, East Asian, Southeast Asian, Indian and Middle Eastern.
In our work, we ignore gender annotations, and use age annotation as a target label and race annotation as group information. 
To act as a reminder that these are the annotations of external labelers, we use the terms such as ``black-annotated'' rather than Black, and ``white-annotated'' rather than white, throughout this work.

\subsection{Model}
We use a standard, randomly-initialized (i.e. not pretrained) ResNet-18 model, as used in our other experiments.
Given an image of a face, the model is trained to perform age classification into one of the nine perceived age buckets provided as part of the FairFace dataset.
To perform multi-class classification, we apply a softmax layer to the model outputs and train with cross-entropy loss.
All other hyperparameters are as described in \cref{sec:app-model-details}.

\section{Calculating amplification factor with linear regression}\label{sec:app-amplification-factor}

Given a vector of estimated accuracy disparities \(\tilde{\textbf{d}}\), and a vector of observed accuracy disparities \(\textbf{d}\), we estimate the amplification factor using OLS linear regression.

In our synthetic setup, we additionally control for the effect of confounds including within-class separability (e.g.\ \emph{aquatic mammal} / \emph{medium mammal} in \cref{fig:cifar-results}{a}), and diagonal separability (e.g.\ \emph{medium mammal} / \emph{flower}), by including them as nuissance regressors.
Our full model is of the form:
\begin{align}
    \textbf{d} = X \beta + \epsilon, \quad
    X = \begin{pmatrix}
        \tilde{\textbf{d}} \\
        \textbf{s}_{a,0}^{a,1} \\
        \textbf{s}_{b,0}^{b,1} \\
        \textbf{s}_{a,0}^{b,1} \\
        \textbf{s}_{a,1}^{b,0}
    \end{pmatrix} \ ,
\end{align}
where amplification factor is the first parameter, \(k = \beta_0\), and \(\textbf{s}_{a,0}^{b,1}\) is the separability (i.e.\ accuracy) between group \(a\), label 0 and group \(b\), label 1.
We use Python \emph{statsmodels} \citep{Seabold2010} to fit the model.

\newpage

\begin{figure*}[h]
    \centering
    \subfloat[Train accuracy]{
        \includegraphics[trim={0 0 0 0},clip,width=0.24\textwidth]{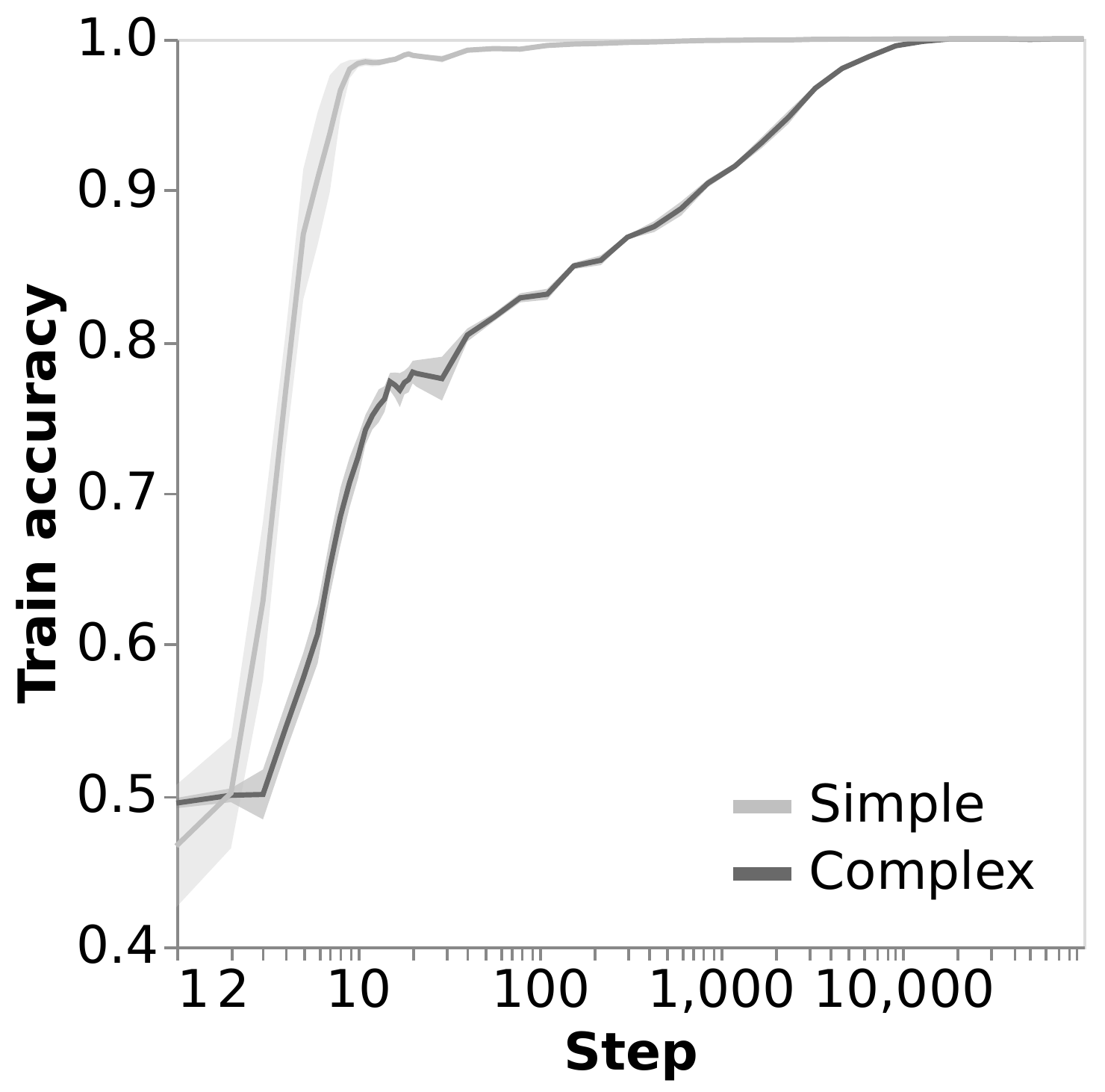}
    }\hspace{20pt}
    \subfloat[Test accuracy]{
        \includegraphics[trim={0 0 0 0},clip,width=0.24\textwidth]{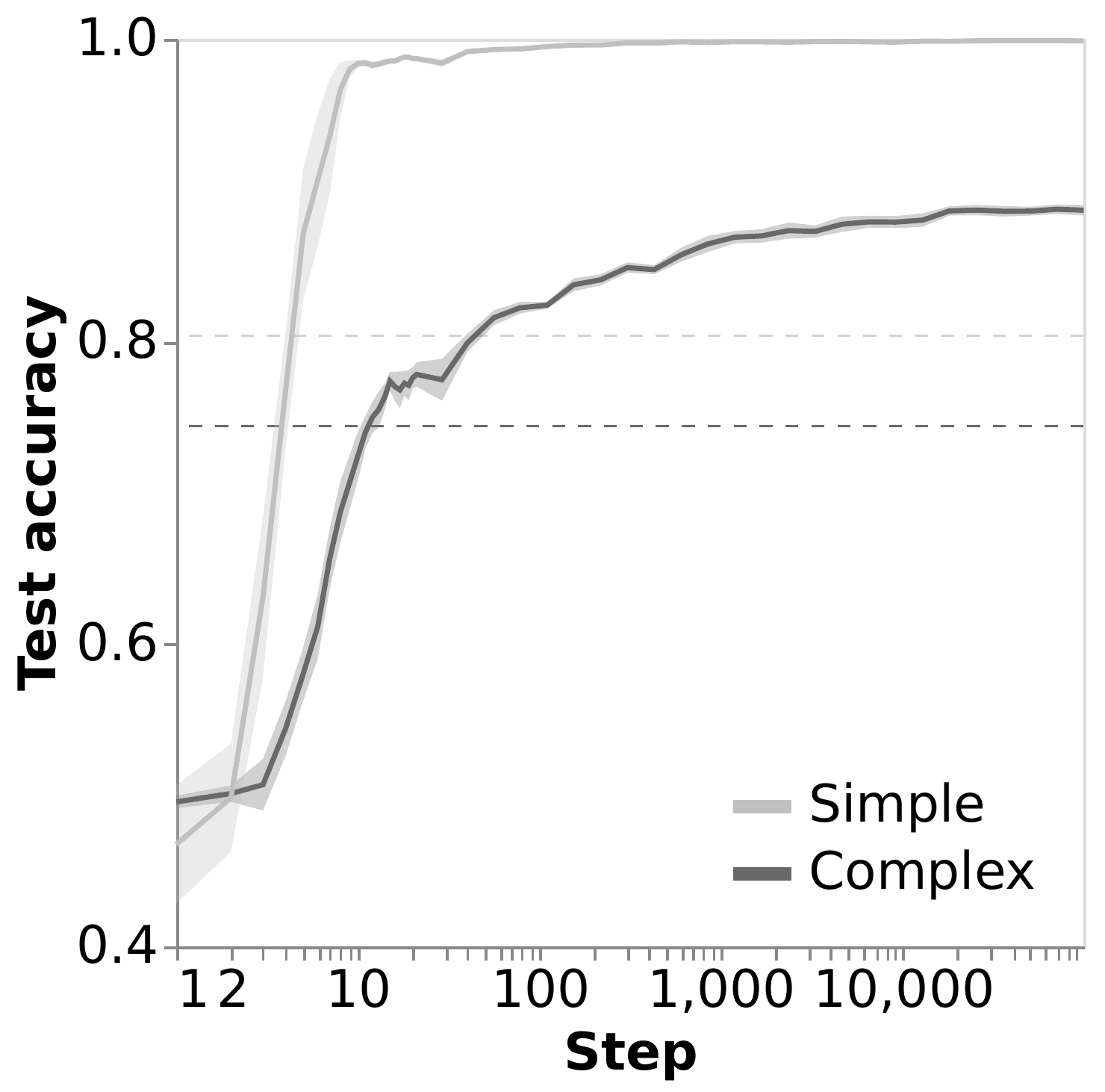}
    }\hspace{20pt}
    \subfloat[Observed disparity]{
        \includegraphics[trim={0 0 0 0},clip,width=0.24\textwidth]{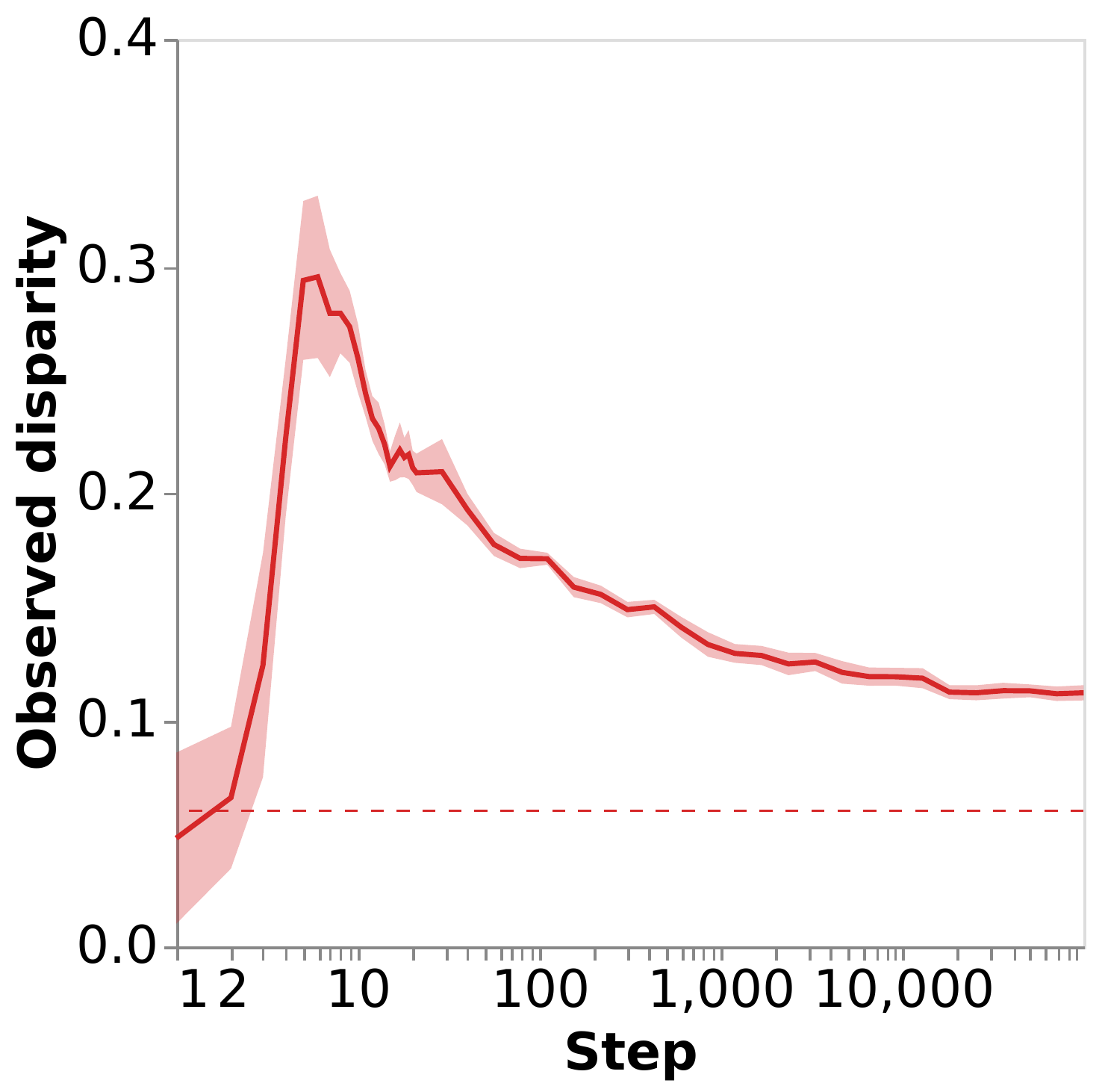}
    }
    \caption{Accuracy and disparity for ResNet-18 on Fashion MNIST \citep{Xiao2017}. Experiment design is identical to \cref{fig:cifar-results}. Simple group is \emph{Trouser}/\emph{Sneaker}; complex group is \emph{T-Shirt}/\emph{Shirt}. Results align with those observed on CIFAR-100.
    \textbf{(a)}~Train accuracy for complex group is learned slower but both groups reach perfect train accuracy. \textbf{(b)}~However, test accuracy for complex group is persistently lower. Dashed lines are binary accuracy from single-group training. \textbf{(c)}~Observed disparity \(d\) peaks early in training but large gap persists after training. Red dashed line is estimated disparity \(\tilde{d}\): observed disparity above this line indicates \emph{amplification}. Shaded area is standard error over 10 runs each with different train/test split.}
    \label{fig:fashion-mnist-results}
\end{figure*}

\begin{figure*}[h]
    \centering
    \subfloat[Train accuracy]{
        \includegraphics[trim={0 0 0 0},clip,width=0.24\textwidth]{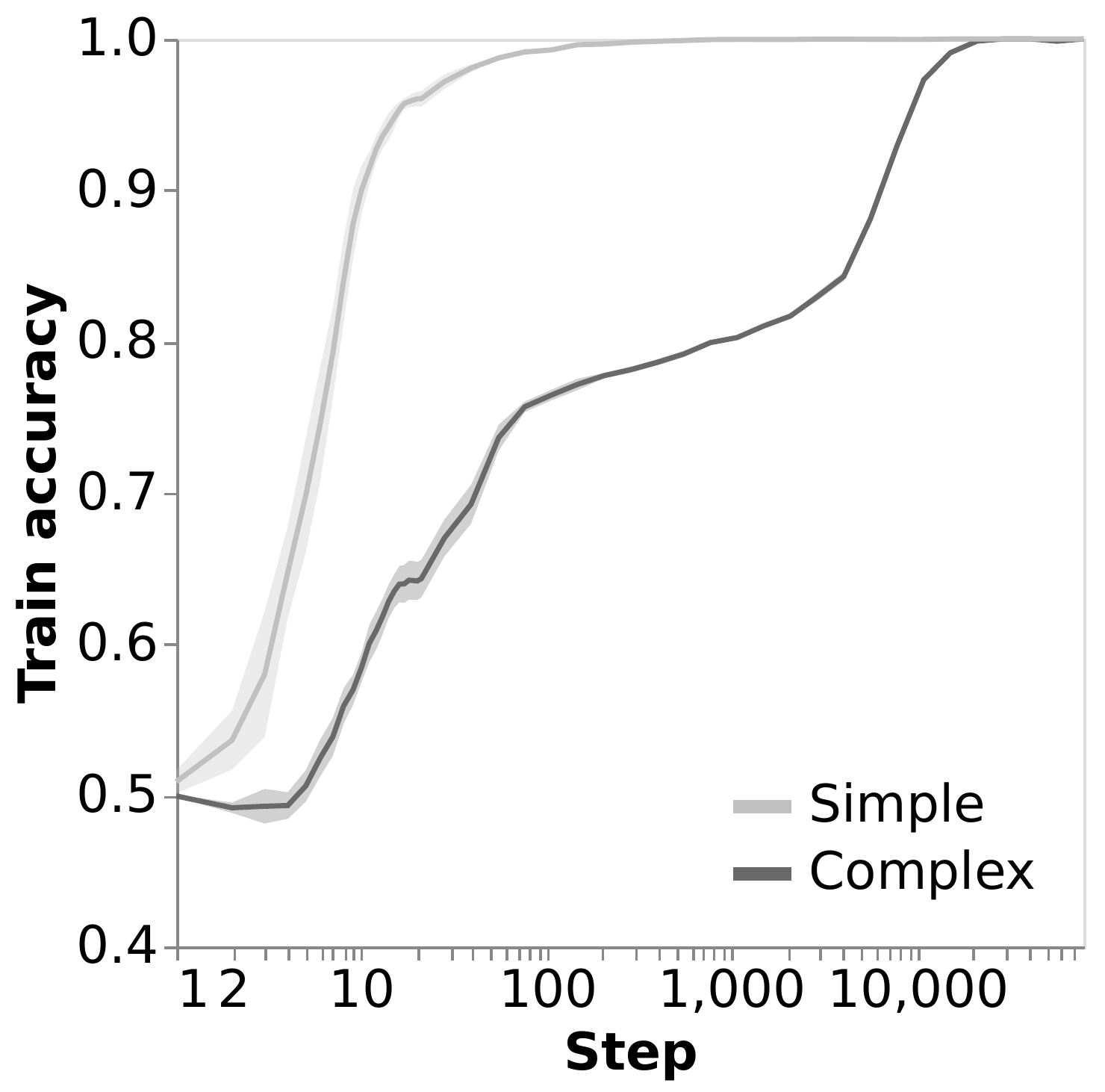}
    }\hspace{20pt}
    \subfloat[Test accuracy]{
        \includegraphics[trim={0 0 0 0},clip,width=0.24\textwidth]{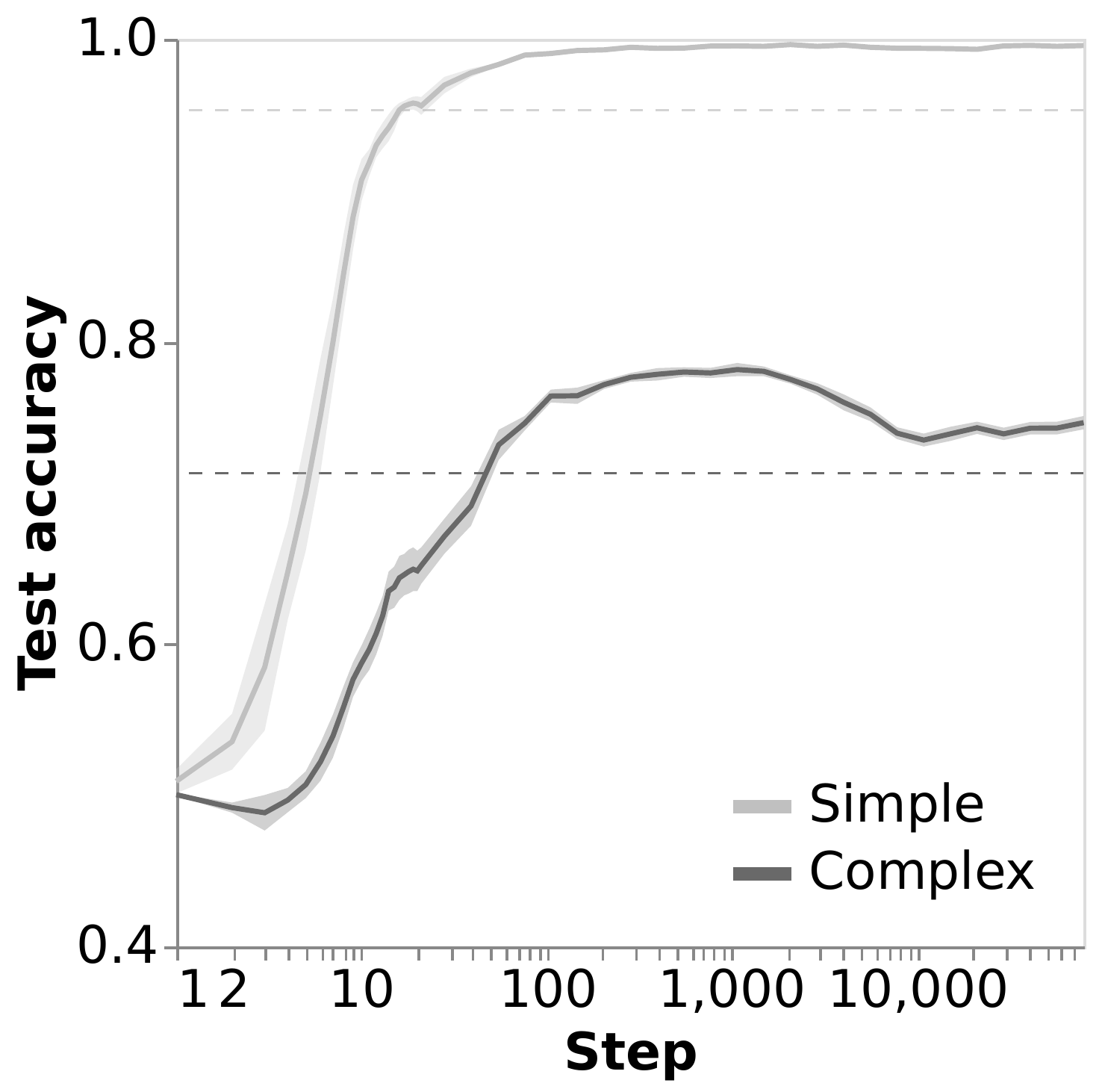}
    }\hspace{20pt}
    \subfloat[Observed disparity]{
        \includegraphics[trim={0 0 0 0},clip,width=0.24\textwidth]{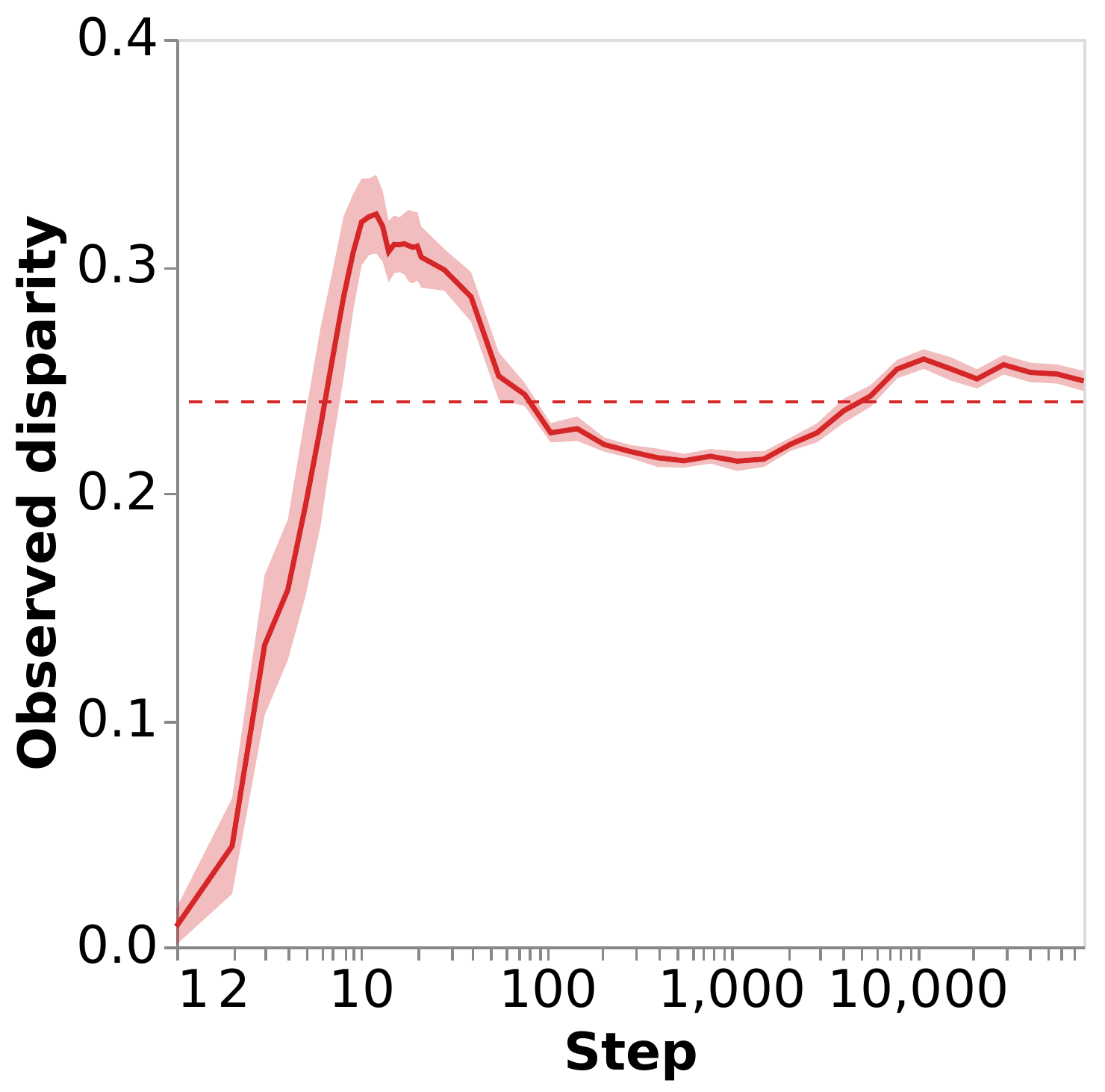}
    }
    \caption{Accuracy and disparity for ResNet-18 on EMNIST letters \citep{Cohen2017}. Experiment design is identical to \cref{fig:cifar-results}. Simple group is \emph{Q}/\emph{X}; complex group is \emph{I}/\emph{L}. Results align with those observed on CIFAR-100.
    \textbf{(a)}~Train accuracy for complex group is learned slower but both groups reach perfect train accuracy. \textbf{(b)}~However, test accuracy for complex group is persistently lower. Dashed lines are binary accuracy from single-group training. \textbf{(c)}~Observed disparity \(d\) peaks early, drops below estimated disparity during training, and then stabilizes at slight amplification.}
    \label{fig:emnist-results}
\end{figure*}

\begin{figure*}[h]
    \centering
    \subfloat[Full CIFAR-100 coarse]{
        \includegraphics[trim={0 0 0 0},clip,width=0.35\textwidth]{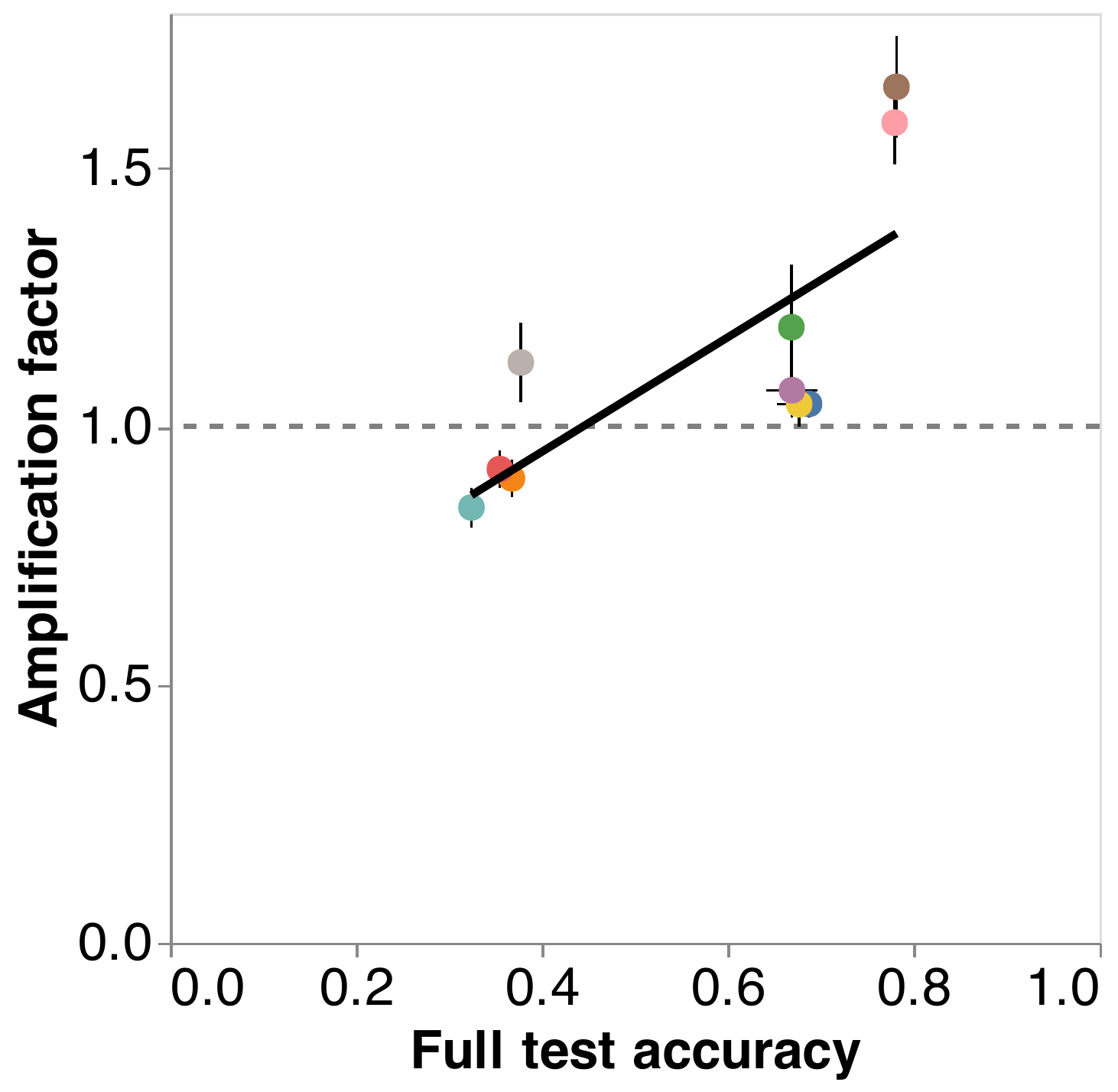}
    }\hspace{20pt}
    \subfloat[Sampled tasks]{
        \includegraphics[trim={0 0 0 0},clip,width=0.35\textwidth]{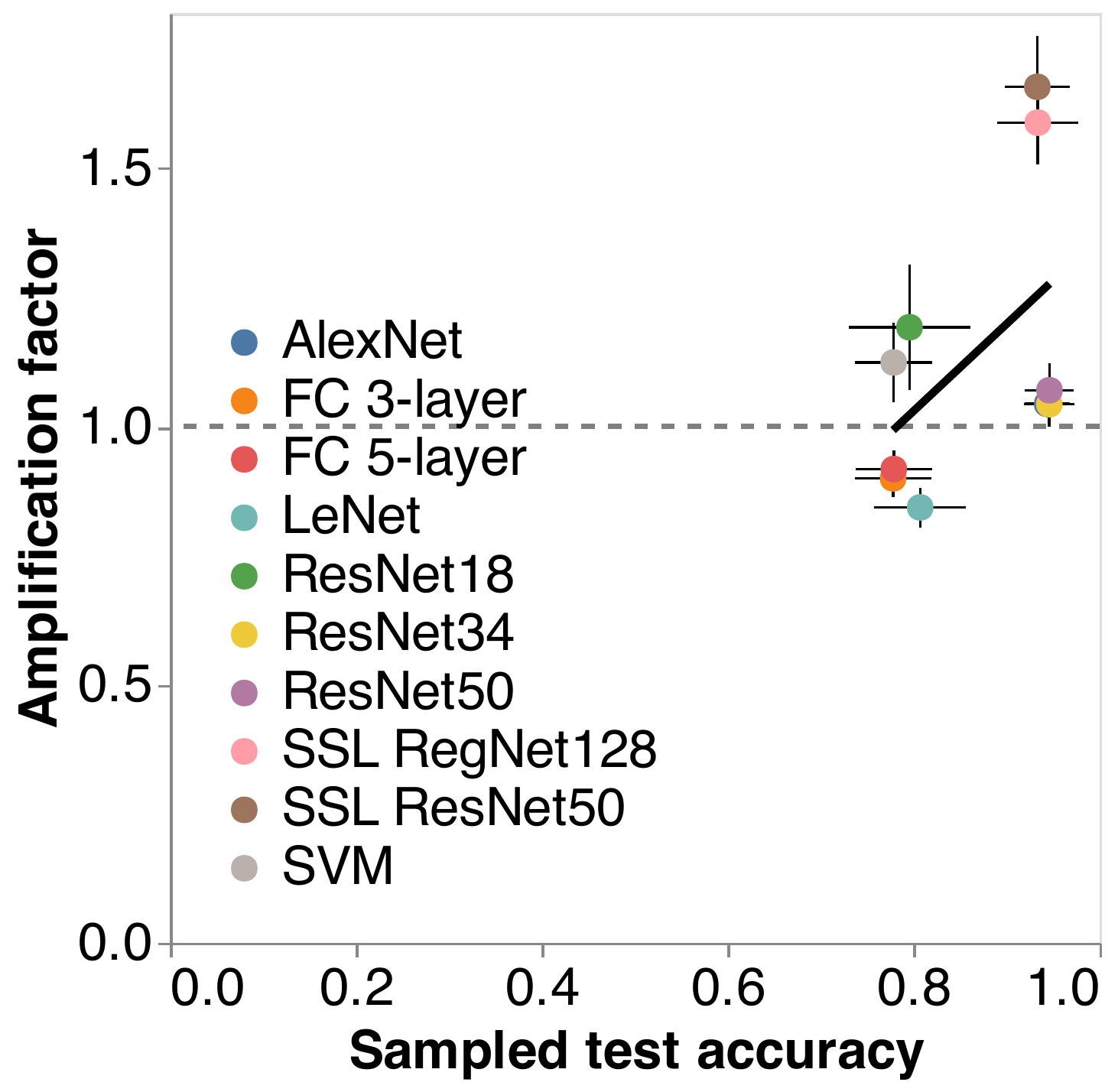}
    }
    \caption{Amplification factor \(k\) as function of average test accuracy on \textbf{(a)} the full CIFAR-100 coarse dataset, and \textbf{(b)} the 30 sampled tasks used for computing amplification factor. \textbf{Choosing a higher accuracy model, e.g./ an SSL model, would increase amplification.} Vertical bars are standard error of the coefficient \(k\). Horizontal bars (barely visible in left panel) are standard deviation of test accuracy over (a) 10 seeds and (b) 10 seeds and 30 tasks. Solid black line fit with linear regression. Dashed gray line is \(k=1\).}
    \label{fig:app-amplification-by-test-acc}
\end{figure*}

\begin{figure*}[h]
    \centering
    \includegraphics[trim={0 0 0 0},clip,width=0.35\textwidth]{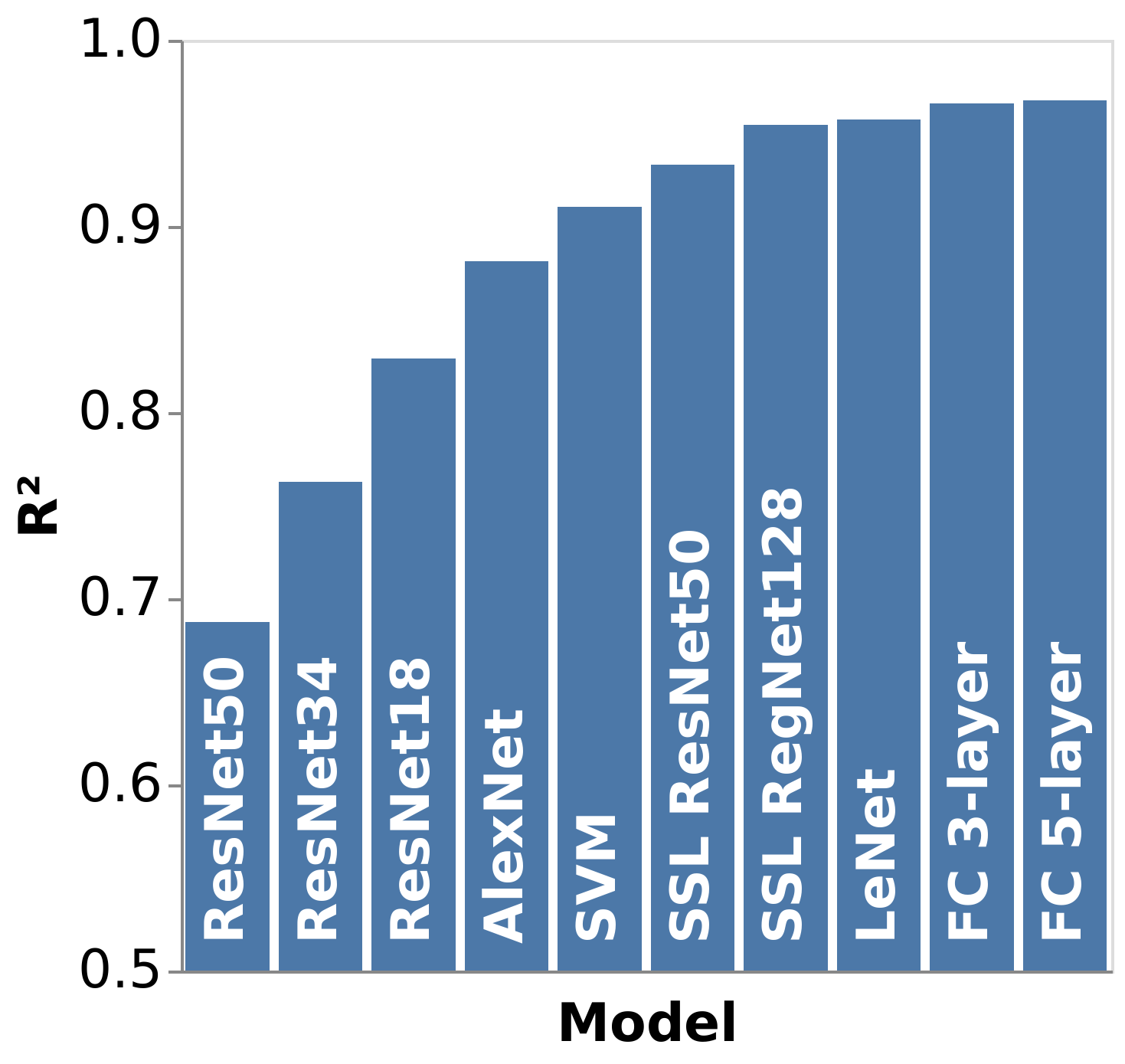}
    \caption{\(R^2\) values for linear regression calculation of amplification factor, for various models, corresponding to \cref{fig:amplification}}
    \label{fig:app-amplification-r2}
\end{figure*}

\begin{figure*}[h]
    \centering
    \subfloat[Width]{
        \includegraphics[trim={0 0 0 0},clip,width=0.23\textwidth]{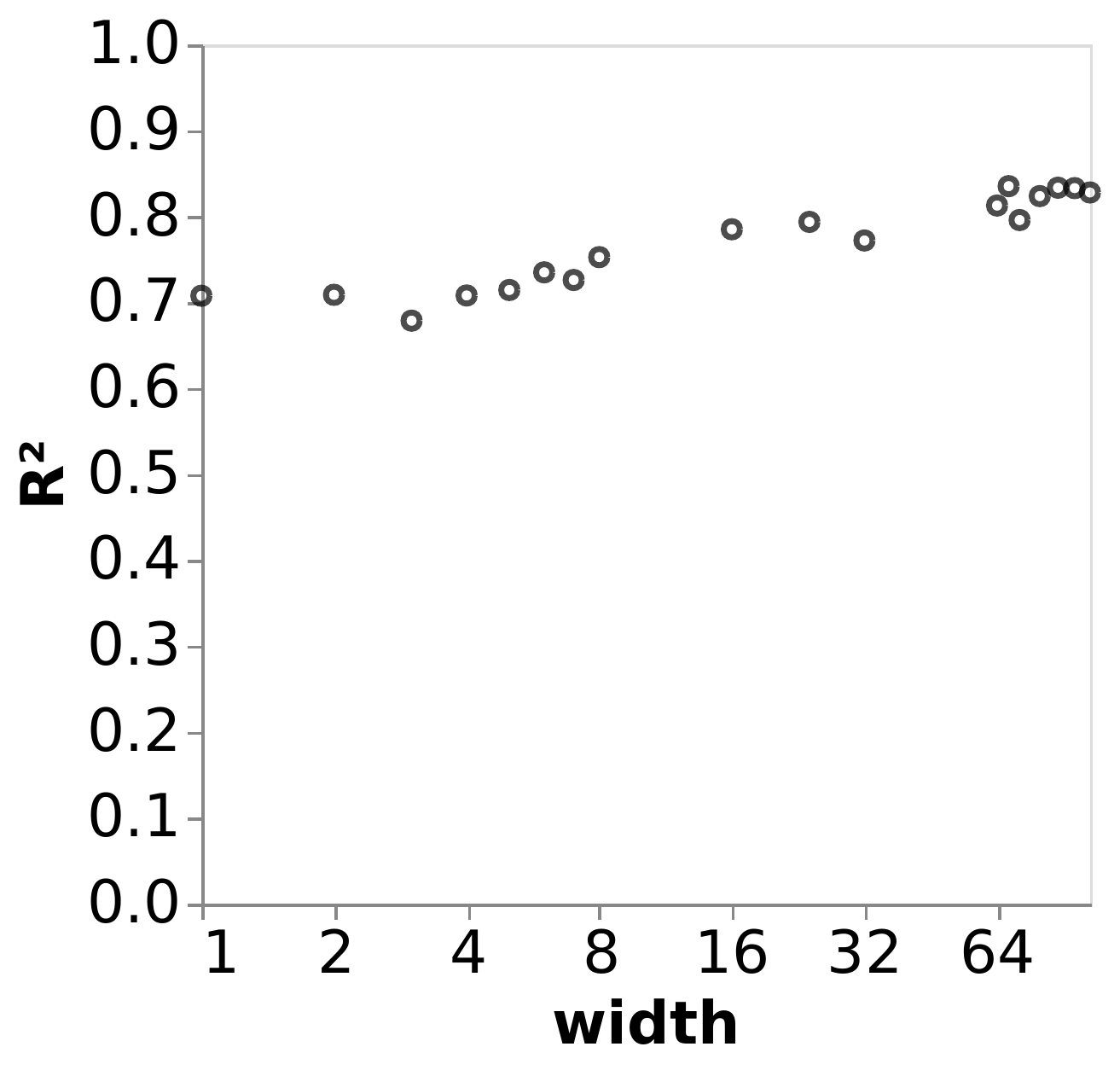}
    }
    \subfloat[Step]{
        \includegraphics[trim={0 0 0 0},clip,width=0.23\textwidth]{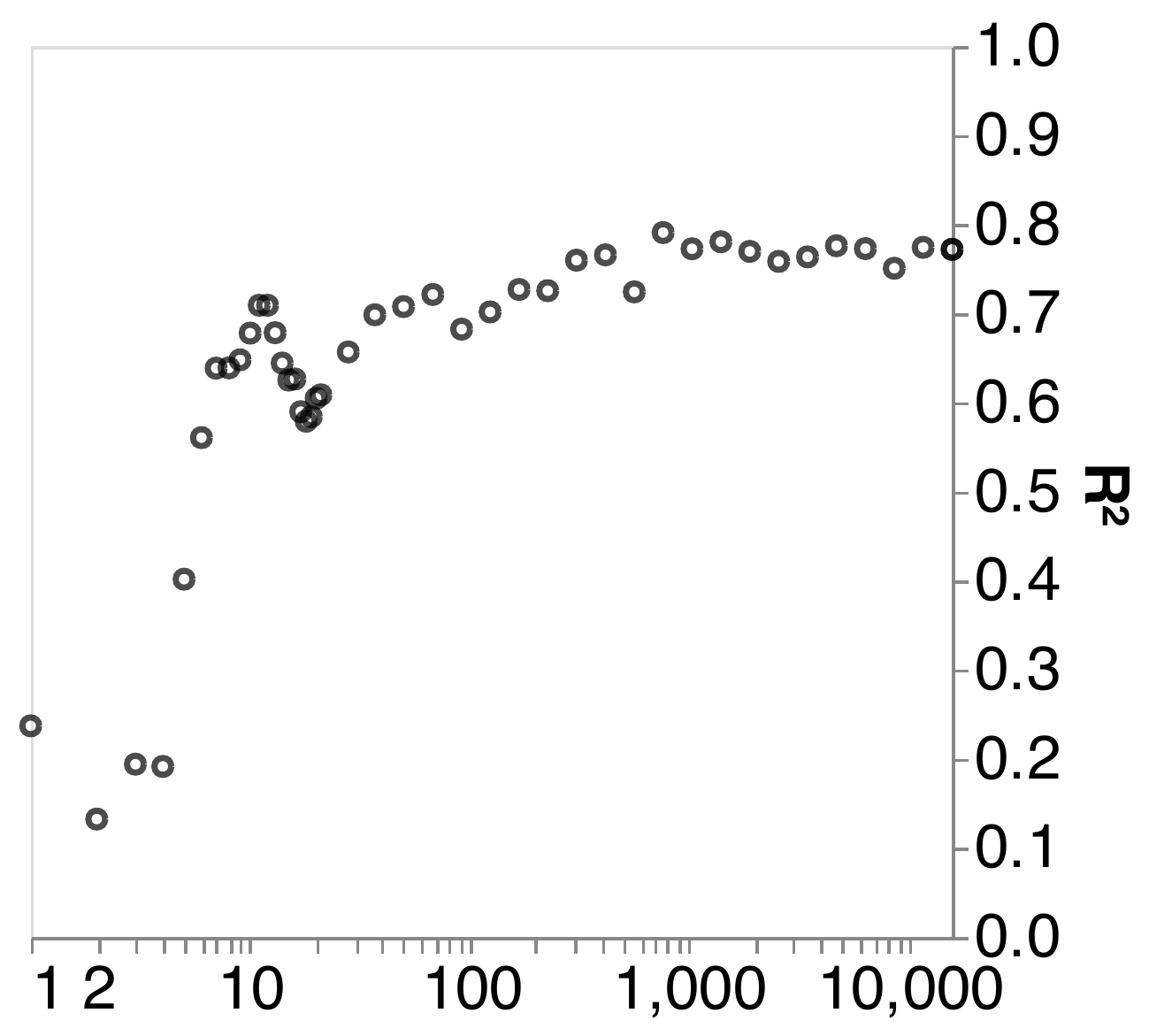}
    }
    \subfloat[Weight decay]{
        \includegraphics[trim={0 0 0 0},clip,width=0.23\textwidth]{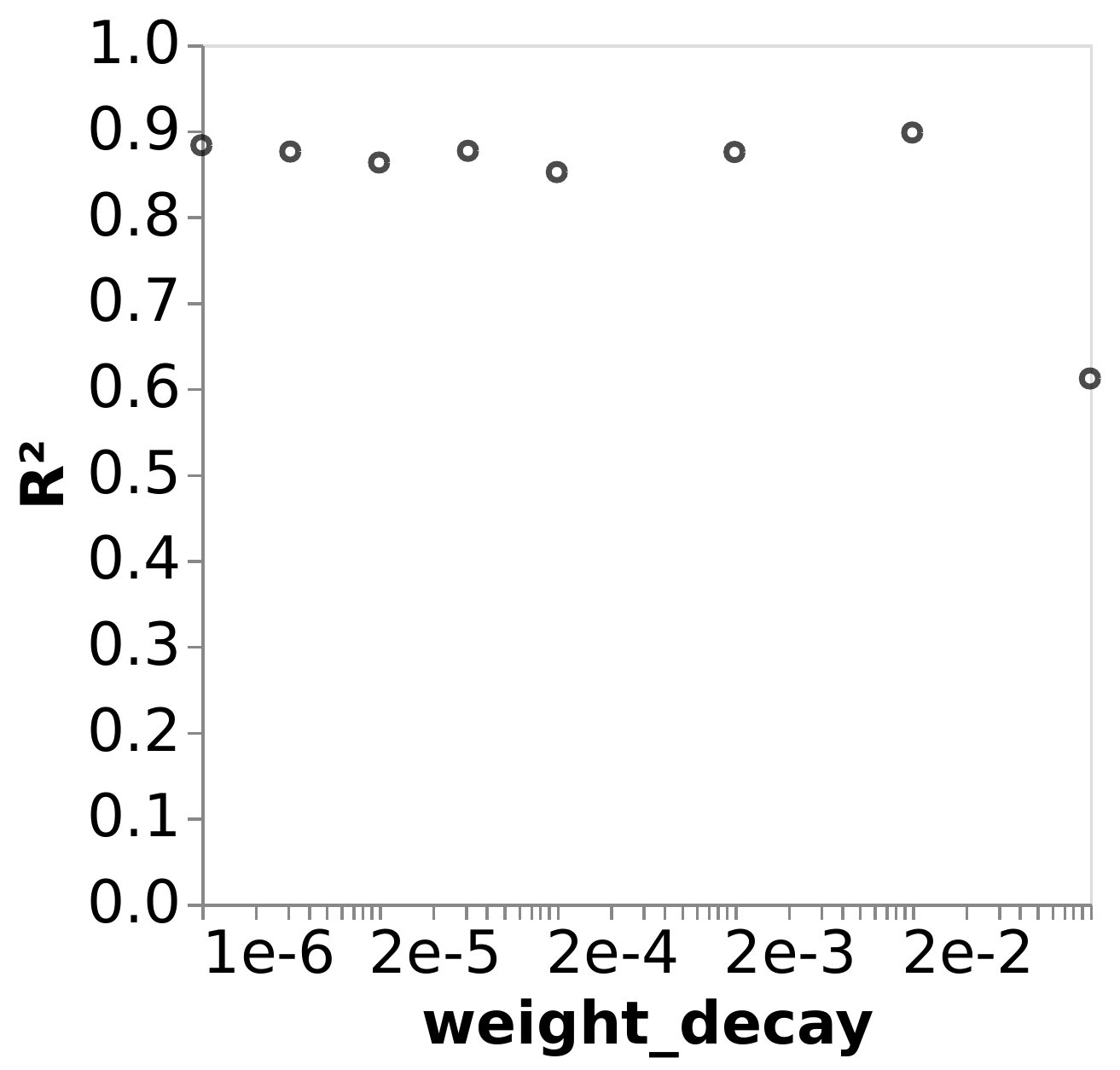}
    }
    \subfloat[Gradient penalty]{
        \includegraphics[trim={0 0 0 0},clip,width=0.23\textwidth]{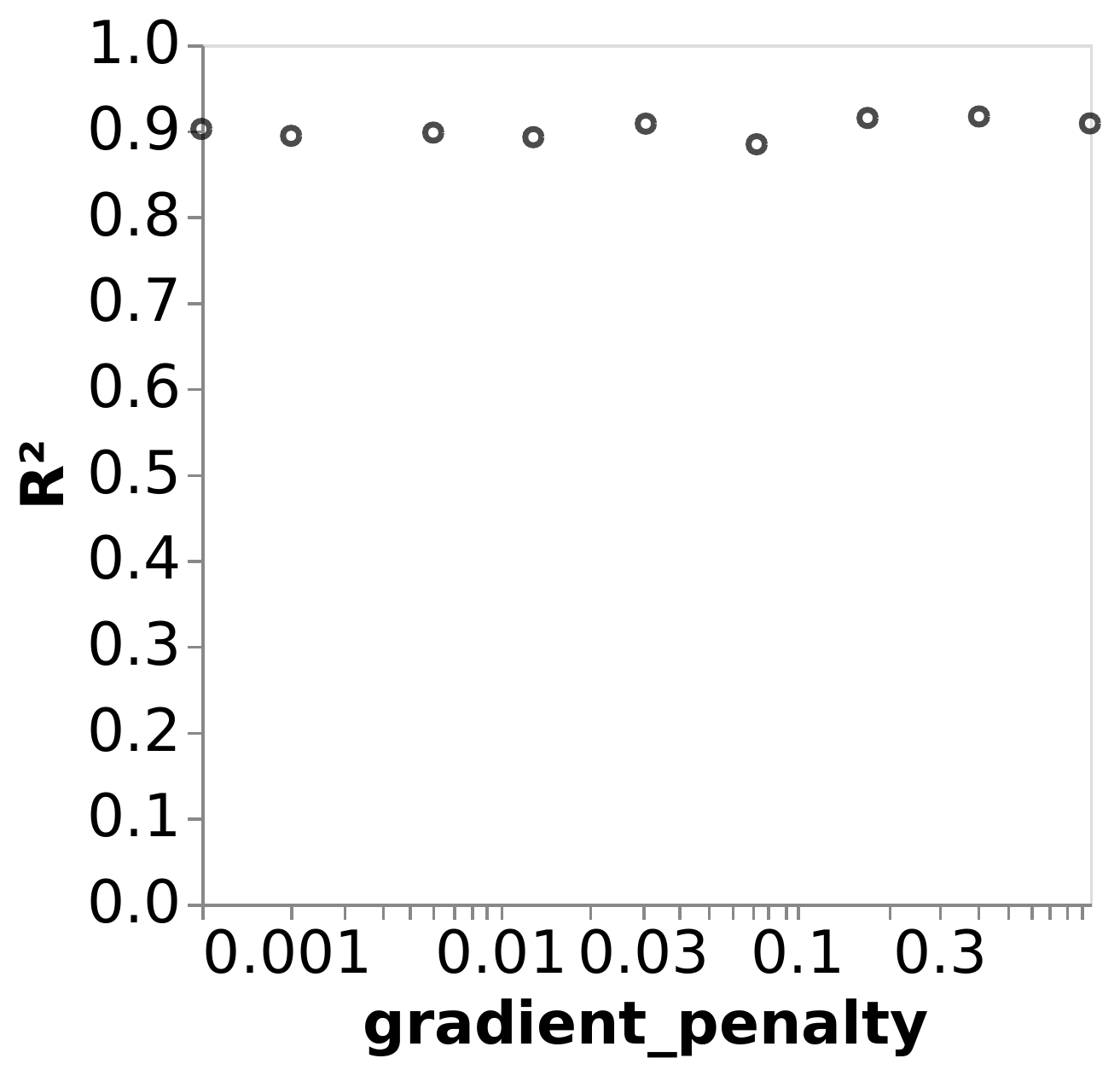}
    }
    \caption{\(R^2\) values for linear regression calculation of amplification factor, corresponding to \cref{fig:design-decisions}.}
    \label{fig:app-design-decisions-r2}
\end{figure*}

\end{document}